\documentclass[10pt,twocolumn,letterpaper]{article}

\usepackage[pagenumbers]{cvpr} % To force page numbers, e.g. for an arXiv version

\usepackage{graphicx}
\usepackage{amsmath}
\usepackage{amssymb}
\usepackage{booktabs}
\usepackage{xcolor}
\usepackage{color, colortbl}
\usepackage{verbatim}

\usepackage{algorithm}
\usepackage{algorithmic}
\usepackage{tabularx}
\usepackage{multirow}
\usepackage{multicol}
\usepackage{bm}
\usepackage{ulem} 
\usepackage[titletoc]{appendix}

\newcommand{\name}{BEVHeight}
\newcommand{\mypara}[1]{\vspace{1mm}\noindent\textbf{#1}}

\usepackage[pagebackref,breaklinks,colorlinks]{hyperref}

\usepackage[capitalize]{cleveref}
\crefname{section}{Sec.}{Secs.}
\Crefname{section}{Section}{Sections}
\Crefname{table}{Table}{Tables}
\crefname{table}{Tab.}{Tabs.}

\def\cvprPaperID{8460} % *** Enter the CVPR Paper ID here
\def\confName{CVPR}
\def\confYear{2023}

\begin{document}

\title{BEVHeight: 
% Towards 
A Robust
% Extrinsic Parameter Free\Tao{Free?} 
Framework for Vision-based Roadside \\
3D Object Detection 
% for  Perception
}

\author{
Lei Yang\textsuperscript{1}\footnotemark[1],  Kaicheng Yu\textsuperscript{2},  Tao Tang\textsuperscript{3},  Jun Li\textsuperscript{1}, Kun Yuan\textsuperscript{4},  Li Wang\textsuperscript{1}, Xinyu Zhang\textsuperscript{1}\footnotemark[2], Peng Chen\textsuperscript{2} \\
	\textsuperscript{1}State Key Laboratory of Automotive Safety and Energy,  Tsinghua University \\
	\textsuperscript{2}Autonomous Driving Lab, Alibaba Group;
 \textsuperscript{3}Shenzhen Campus, Sun Yat-sen University \\
 \textsuperscript{4}Center for Machine Learning Research, Peking University \\ 
    \begin{normalsize}${\tt \{yanglei20@mails, lijun1958@, xyzhang@, wangli\_thu@mail\}.tsinghua.edu.cn}$\end{normalsize} \\
    \begin{normalsize}${\tt \{kaicheng.yu.yt, trent.tangtao\}@gmail.com; kunyuan@pku.edu.cn; yuanshang.cp@alibaba\mbox{-}inc.com}$\end{normalsize}}

% \and

% \maketitle
%figure1

\twocolumn[{%
\renewcommand\twocolumn[1][]{#1}%
\maketitle
\begin{center}
\centering
\captionsetup{type=figure}
\vspace{-0.38cm}
\begin{center}
\includegraphics[width=0.99\textwidth]{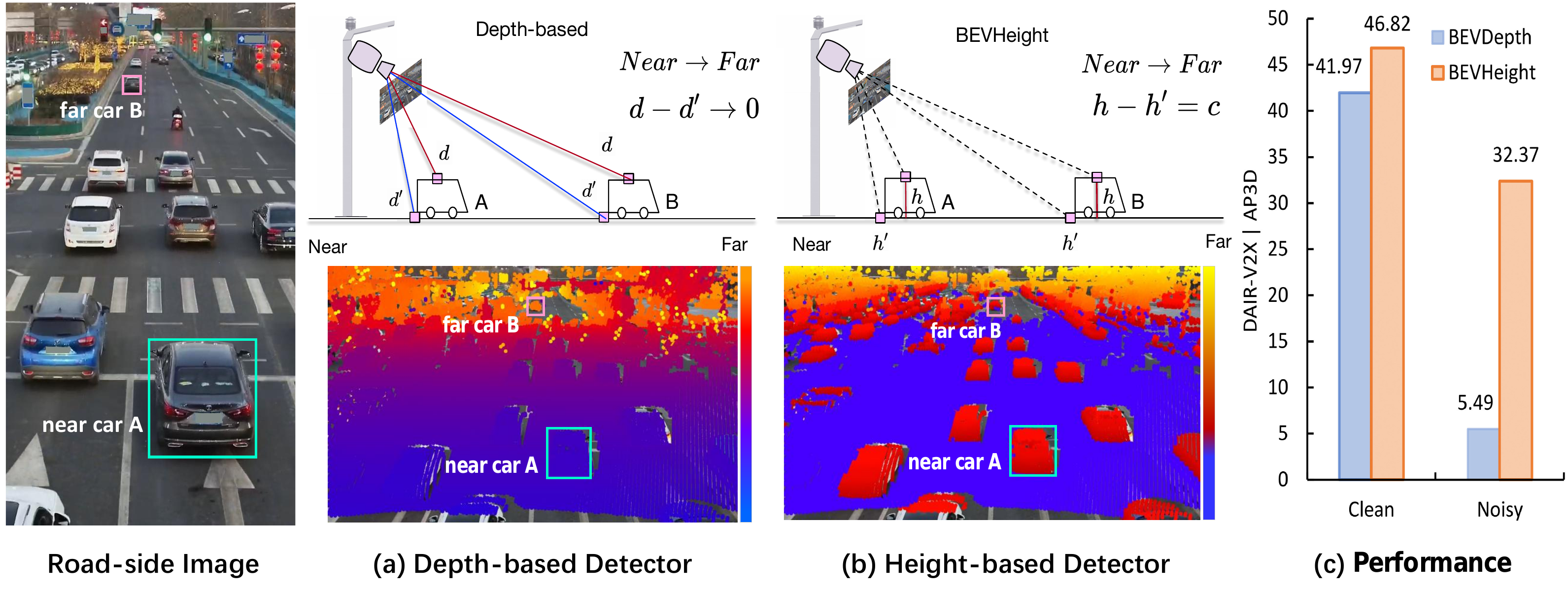}
\end{center}
\setlength{\belowcaptionskip}{-0.005cm} 
\captionof{figure}{
\textbf{(a)} To produce 3D bounding boxes out of a monocular image, state-of-the-art methods firstly predict the per-pixel depth either explicitly or implicitly to determine the 3D location of foreground objects with the background. However, when we plot the per-pixel depth on the image, we notice that the differences between points on the car roof and surrounding ground quickly shrink when the car moves away from the camera, making it sub-optimal to optimize especially for far objects. \textbf{(b)} On the contrary, we plot the per-pixel height to the ground and observe that such difference remains agnostic regardless of the distance, and visually is superior for the network to detect objects. However, one cannot directly regress the 3D location by solely predicting the height.  \textbf{(c)} To this end, we propose a novel framework, \name{} to address this issue. Empirical results reveal that our method surpasses the best method by a margin of 4.85\% on clean settings and over 26.88\% on noisy settings.
% Without loss of generality, we mark two point, one on the car roof, another on the nearest ground plane, and denote the depth of camera center to these points as $d$ and $d'$. 
% \textbf{(a)} We visualize per-pixel depth and height of road-side cameras. State-of-the-art  To effectively differentiate the point on the vehicle and nearest ground, 
% % state-of-the-art methods rely on depth as a supervision, i.e. to first predict the depth then recover the 3D geometry. 
% state-of-the-art methods rely on implicitly or explicity depth, i.e. to first predict the depth then recover the 3D geometry. 
% However, the depth difference will drastically shrink to zero when the distance between the vehicle and camera increases. This potentially is sub-optimal to . Motivated by the residual learning regime, we propose a novel framework that predicts height instead of depth, where the difference between car and ground remains unchanged regardless the distance. As analyzed later, our framework significantly ease the optimization process and is more robust. \textbf{(c)} 
}
\label{fig:teaser}
\end{center}%
}]

% \begin{figure*}[!t]
% 	\centering
% 	\includegraphics[width=\textwidth]{BEV-Height/figures/intro.pdf}
% 	\caption{\textbf{(a)} Visualize per-pixel depth and height of road-side cameras. \textbf{(b)} To effectively differentiate the point on the vehicle and nearest ground, state-of-the-art methods rely on depth as a supervision, i.e. to first predict the depth then recover the 3D geometry. However, the depth difference will drastically shrink to zero when the distance between the vehicle and camera increases, thus is sub-optimal to learn. On the contrary, we propose a novel framework to regress height, where the difference between car and ground remains agnostic to the distance and is more robust. \textbf{(c)} Empirical results reveal that our method surpass previous best method by a margin of 3\% on clean setting and over 20\% on noisy settings. }
% \label{fig:teaser}
% \end{figure*}

% %figure1
% \begin{figure}[t]
% 	\centering
% 	\includegraphics[width=8.5cm]{BEV-Height/figures/Teaser.pdf}
% 	\caption{\textbf{The comparison of depth and height distribution.}}
% \label{fig:one}
% \end{figure}

% %figure2
% \begin{figure}[htbp]
% 	\centering
% 	\includegraphics[width=8.5cm]{BEV-Height/figures/perspective_principle.pdf}
% 	\caption{\textbf{The comparison of depth and height distribution.}}
% \label{fig:two}
% \end{figure}

\renewcommand{\thefootnote}{\fnsymbol{footnote}}
\footnotetext[1]{Work done during an internship at DAMO Academy, Alibaba Group.}
\footnotetext[2]{Corresponding Author.}
\renewcommand{\thefootnote}{\arabic{footnote}}

%%%%%%%%% ABSTRACT
\begin{abstract}
% Perception is one of the most critical tasks in autonomous driving, where the mainstream methods only rely on ego-vehicle sensors, such as cameras, to recognize the surrounding worlds. However, these sensors are limited to seeing objects within a fixed radius of the vehicle center. With the fast growth of intelligent infrastructures, roadside camera units have the potential to see the world beyond the visual range. 
While most recent autonomous driving system focuses on developing perception methods on ego-vehicle sensors, people tend to overlook an alternative approach to leverage intelligent roadside cameras to extend the perception ability beyond the visual range. 
% However, unlike ego-vehicle cameras, these roadside ones suffer from various challenges. For example, the installation position and camera poses are usually inconsistent even on the same road, which greatly challenges the robustness of the perception method. 
We discover that the state-of-the-art vision-centric bird's eye view detection methods have inferior performances on roadside cameras. 
This is because these methods mainly focus on recovering the depth regarding the camera center, where the depth difference between the car and the ground quickly shrinks while the distance increases. In this paper, we propose a simple yet effective approach, dubbed BEVHeight, to address this issue. In essence, instead of predicting the pixel-wise depth, we regress the height to the ground to achieve a distance-agnostic formulation to ease the optimization process of camera-only perception methods. On popular 3D detection benchmarks of roadside cameras, our method surpasses all previous vision-centric methods by a significant margin. The code is available at {\url{https://github.com/ADLab-AutoDrive/BEVHeight}}.
% \href{https://github.com/ADLab-AutoDrive/BEVHeight}{here}.

\comment{
   Overlooked roadside perception has great potential to solve the issues of over-occlusion and long-range perception, which is believed to facilitate more intelligent and safer autonomous driving. Vision-centric BEV perception perceives inherent merits in fusioning sensors from adjacent road side units (RSU). However, the existing BEV perception methods mainly aim at autonomous driving and are dedicated to achieving higher accuracy. The generalization challenge caused by various cameras with ambiguous mounting positions and inconsistent intrinsic parameters on RSU is disregarded. In this paper, we find that the height information perpendicular to the ground has three benefits from the perspective of road side unit: 1) Centralized distribution is conducive to improving accuracy 2) Not depending on perspective principle makes it robust to intrinsic perturbation, 3) Weakly correlated with image area make it not easily affected by ambiguous mounting viewpoint. Motivated by the conclusions above, we propose a novel height-based BEV framework for roadside perception, dubbed BEVHeight, whose PV-BEV transformation relies on the implicit height estimation, thus achieving higher accuracy and stronger generalization on the roadside. Experiments on the Rope3D and DAIR-V2X 3D object detection benchmark show that our framework surpasses the state-of-the-art methods under the standard settings. Under the camera's internal and external parameters disturbed setting, our method yields the best robustness.
  }
\end{abstract}
% 对于自动驾驶，感知很重要，现在关注于车辆自身的感知，忽视了路端的优势（遮挡/远距离）；（为什么路端）
% 最近新兴的车端BEV感知在路端也可以发挥他的优点（传感器融合）；（为什么BEV）
% 但是路端传感器的复杂环境（安装位置/内外参类型）对流行的视觉BEV感知方法的精度和泛化性带来了挑战；（有什么问题）
% 我们发现了问题在于现在的最好的sota BEV方法依赖于准确的depth估计，但是在路端depth分布差异大/传感器内外参差异大（问题）
% 于是我们提出了BEV-height，巧妙的选择了height估计，对比depth，其优点（height分布集中/透视原理？）（解决问题）
% 取得了很好的精度和泛化性结果（结果）

%%%%%%%%% BODY TEXT
% \vspace{-0.3cm}
\section{Introduction}
\label{sec:intro}
The rising tide of autonomous driving vehicles draws vast research attention to many 3D perception tasks, of which 3D object detection plays a critical role.  
While most recent works tend to only rely on ego-vehicle sensors, there are certain downsides of this line of work that hinders the perception capability under given scenarios. For example, as the mounting position of cameras is relatively close to the ground, obstacles can be easily occluded by other vehicles to cause severe crash damage. To this end, people have started to develop perception systems that leverage intelligent units on the roadside, such as cameras, to address such occlusion issue and enlarge  perception range so to increase the response time in case of danger \cite{ye2022rope3d, yu2022dair, song2022efficient, huang2022rd,cui2022coopernaut, xu2022v2x}. To facilitate future research, there are two large-scale benchmark datasets~\cite{yu2022dair, ye2022rope3d} of various roadside cameras and provide an evaluation of certain baseline methods. 

% \KY{say something about the current vision-centric bird's eye view method} 
% Thanks to the two recently published benchmarks, XX, YY, 
Recently, people discover that, in contrast of directly projecting the 2D images into a 3D space, leveraging a bird's eye view~(BEV) feature space that reduce the Z-axis degree of freedom can significantly improve the perception performance of vision centric system. One line of recent approach, which constitutes the state-of-the-art in camera only method, is to generate implicitly or explicitly the depth for each pixel to ease the optimization process of bounding box regression. However, as shown in \cref{fig:teaser}, we visualize the per-pixel depth of an example roadside image and notice a phenomenon. Consider two points, one on the roof of a car and another on the nearest ground. If we measure the depth of these points to camera center, namely $d$ and $d'$ respectively, the difference between these depth $d-d'$ would drastically decrease when the car moves away from the camera. We conjecture this leads to two potential downsides: i) unlike the autonomous vehicle that has a consistent camera pose, roadside ones usually have different camera pose parameters across the datasets, which makes regressing depth hard; ii) depth prediction is very sensitive to the change of extrinsic parameter, where it happens quite often in the real world. 

On the contrary, we notice that the height to the ground is consistent regardless of the distance between car and camera center. To this end, we propose a novel framework to predict the per-pixel height instead of depth, dubbed \name{}. Specifically, our method firstly predicts categorical height distribution for each pixel to project rich contextual feature information to the appropriate height interval in wedgy voxel space. Followed by a voxel pooling operation and a detection head to get the final output detections. Besides, we propose a hyperparameter-adjustable height sampling strategy. Note that our framework does not depend on explicit supervision like point clouds. 

% \KY{to complete after experiment is finalized} Through extensive experiments on DAIR-V2X and Rope3D, our method achieves xxxx over yyy ....
We conduct extensive experiments on two popular large-scale benchmarks for roadside camera perception, DAIR-V2X~\cite{yu2022dair} and Rope3D~\cite{ye2022rope3d}. On traditional settings where there is no disruption to the cameras, our \name{} achieves the state-of-the-art performance and surpass all previous methods, regardless of monocular 3D detectors or recent bird's eye view methods by a margin of 5\%. In realistic scenarios, the extrinsic parameters of these roadside units can be subject to changes due to various reasons, such as maintenance and wind blows. We simulate these scenarios following \cite{yu2022benchmarking} and observes a severe performance drop of the BEVDepth, from 41.97\% to 5.49\%. Compared to these methods, we showcase the benefit of predicting the height instead of depth and achieve 26.88\% improvement over the BEVDepth~\cite{li2022bevdepth}, which further evidences the robustness of our method.

\section{Related Work}
\mypara{Roadside Perception.}
Concurrent perception efforts for autonomous driving are mainly limited to the ego vehicle~\cite{caesar2020nuscenes, sun2020waymo}. While the roadside perception, which comparatively has a longer perceptual range and more robustness to occlusion and long-time event prediction, is mainly under-explored. Recently, some pioneers have present roadside datasets~\cite{ye2022rope3d, yu2022dair}, hoping to facilitate the 3D perception tasks in roadside scenarios. Compared with the vehicle perceptual system, which only observes surroundings in a short distance, the roadside cameras, mounted on poles a few meters above the ground, can provide long-range perception. However, the cameras mounted on roadside units have ambiguous mounting positions and 
%inconsistent intrinsic 
variable extrinsic parameters, which bring critical challenges to current perception models. In this paper, we take the advances and challenges of roadside cameras into account, and design an efficient and robust roadside perception framework, \name{}.

% \mypara{Vision-based Multi-View BEV perception.}
%  single-camera setting and multi-camera setting
\mypara{Vision Centric BEV Perception.}
Recent vision-centric works predict objects in 3D space, which is very suitable for applying multi-view feature aggregation under BEV for autonomous driving. Popular methods can be divided into transformer-based and depth-based schema. 
Following DETR3D~\cite{wang2022detr3d}, transformer-based detectors design a set of object queries~\cite{liu2022petr, liu2022petrv2, jiang2022polarformer, Chen2022PolarPF, Saha2022TranslatingII} or BEV grid queries\cite{li2022bevformer}, then perform the view transformation through cross-attention between queries and image features. 
Following LSS~\cite{philion2020lift}, depth-based methods ~\cite{reading2021cadnn, huang2021bevdet, huang2022bevdet4d} explicitly predict the depth distribution and use it to construct the 3D volumetric feature. Followup works introduce depth supervision from the LiDAR sensors ~\cite{li2022bevdepth} or multi-view stereo techniques ~\cite{wang2022sts, li2022bevstereo, park2022solofusion} to improve the depth estimation accuracy and achieve state-of-the-art performance. 
Additionally, transformer-based detectors' implicit 3D location information also can benefit from accurate depth cues. Inspired by~\cite{zhang2022monodetr,huang2022monodtr}, CrossDTR~\cite{tseng2022crossdtr} proposed depth-guided transformers, which compose depth-aware embedding from depth maps and are supervised by ground truth depth maps to enhance performance.
However, when applying these methods to roadside perception, the bonus of accurate depth information fades. As the complex mounting positions and variable extrinsic parameters of the roadside cameras, predicting depth from them is difficult. In this work, our \name{} utilizes the height estimation to achieve state-of-the-art performance and best robustness of roadside 3D object detection.

% method
\section{Method}
%figure3
% \begin{figure}[t!]
% 	\centering
% 	\includegraphics[width=8.5cm]{BEV-Height/figures/depth-height1.pdf}
% 	\caption{
% 	\textbf{The comparison of predicting height and depth.} 
% % 	\textbf{The histogram distribution of depth and the height from the ground.} all data are derived from the annotations of DAIR-V2X-I\cite{yu2022dair} dataset. (a) reveals the depth histogram distribution. (b) indicates the height counterpart.
% % 和 Overall Architecture 图有点重复
% 	}
% \label{fig:histogram-depth-height}
% \end{figure}

\begin{figure*}[h!t]
	\centering
	\includegraphics[width=0.8\textwidth]{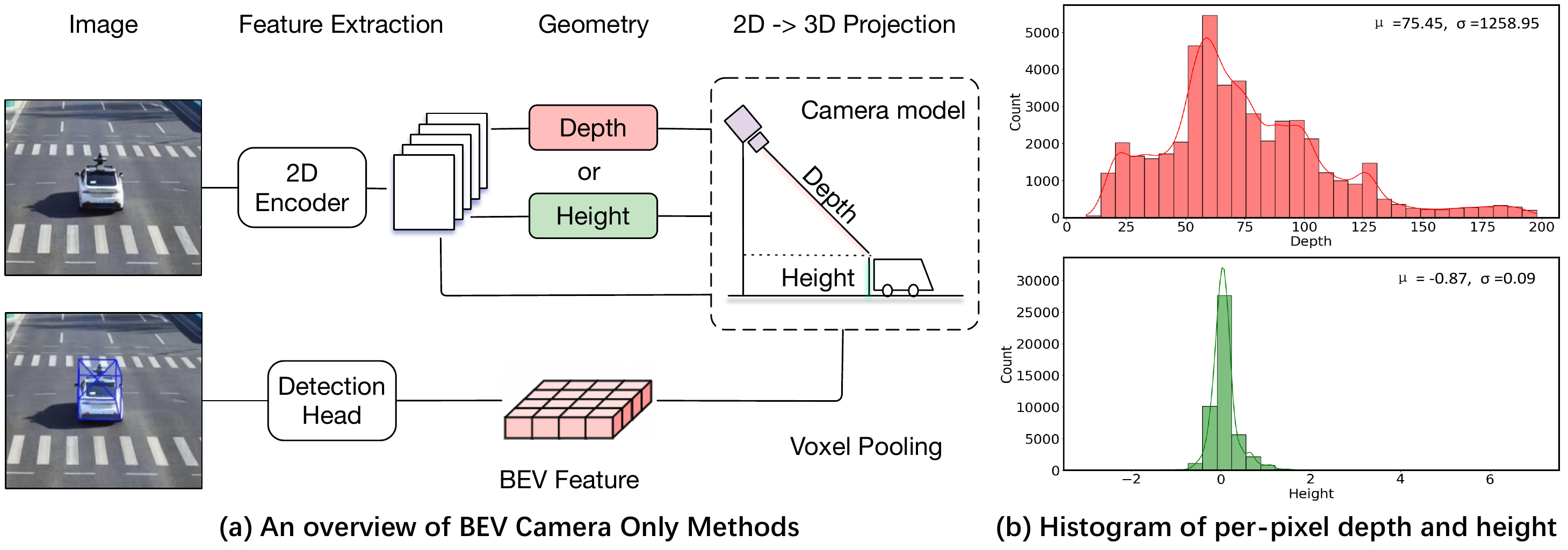}
	\caption{
	\textbf{The comparison of predicting height and depth.} 
	\textbf{(a)} We present the overview of previous depth based monocular 3D detection methods and our proposed \name{}. Note that we propose a novel 2D to 3D projection module. \textbf{(b)} We plot the histogram of per-pixel depth~(top) and ground-height~(bottom). We can clearly observe that the range of depth is over 200 meters while the height is within  5 meters, which makes height much easier to learn.
% 	\textbf{The histogram distribution of depth and the height from the ground.} all data are derived from the annotations of DAIR-V2X-I\cite{yu2022dair} dataset. (a) reveals the depth histogram distribution. (b) indicates the height counterpart.
% 和 Overall Architecture 图有点重复
	}
 \vspace{-0.1cm}
\label{fig:histogram-depth-height}
\end{figure*}

\begin{figure*}[ht]
\centering 
\includegraphics[width=0.8\textwidth]{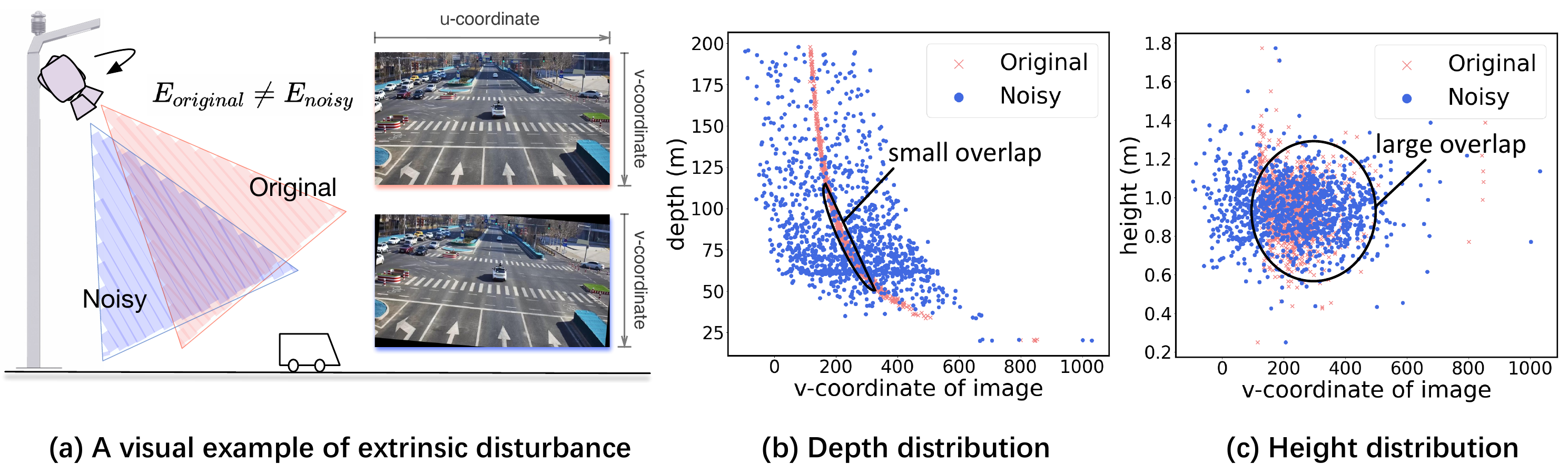}
% 	\caption{\textbf{The correlation between the image row coordinates of the object and the corresponding depth and height.} ``R\&P" implies roll and pitch. The ``disturbed R\&P" represents a rotation offset along roll and pitch directions in normal distribution (a) is the scatter diagram of the depth distribution. (b) is for the height from the ground.
	
% 	the position of the object in the image, which can be defined as $(u,v)$, and $v$ denotes its row coordinate of image
% 	}
\caption{\textbf{The correlation between the object's row coordinates on the image with its depth and height.} The position of the object in the image, which can be defined as $(u,v)$, and $v{\raisebox{0mm}{-}}coordinate$ denotes its row coordinate of the image. (a) A visual example of the noisy setting, adding a rotation offset along roll and pitch directions in the normal distribution. (b) is the scatter diagram of the depth distribution. (c) is for the height from the ground. We can find, compared with depth, the noisy setting of height has larger overlap with its original distribution, which demonstrates height estimation is more robust.
}
\vspace{-0.3cm}
\label{fig:five}
\end{figure*}

% In this section, we introduce our proposed \name{} in detail. 
We first give a brief problem definition of camera-only 3D object detection on roadside. We then analyze the downside of predicting depth that is widely-adopted in current camera-only methods and show the benefit of using height instead. Subsequently, we present our framework in detail. 
% and the generalization assessments under the camera's external parameters disturbed setting in Sec.~\ref{sec:problem_definition}. 
% Then, By analyzing the training data, we justify the superiority of height estimation over the depth in constructing 3D features from the roadside view in Sec.~\ref{sec:delving_into_the_height}. Subsequently, we give an overview of the whole framework and clarify our key contributions including height network and the projection process from 2D to 3D in Sec.~\ref{sec:BEVHeight}.
%figure6
\begin{figure*}[t]
	\centering
	\includegraphics[width=0.90\textwidth]{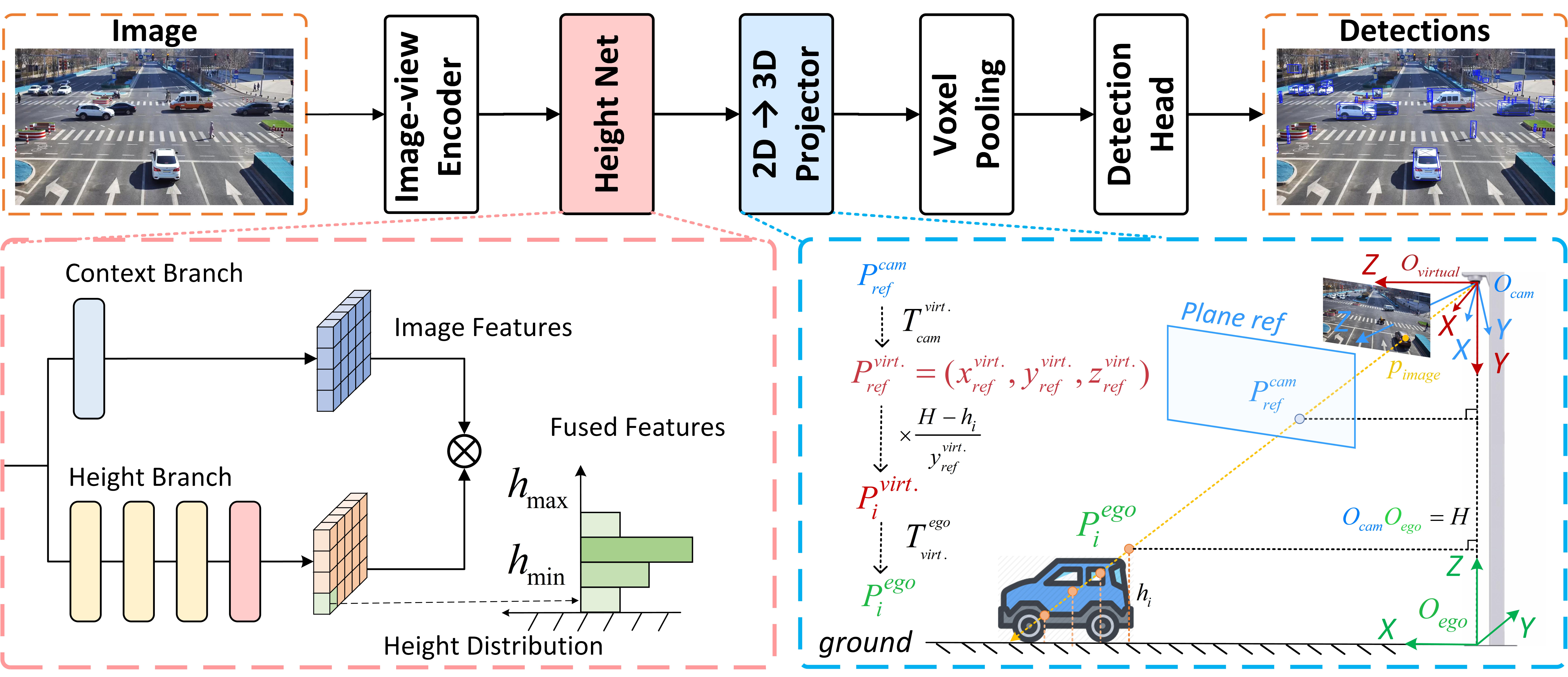}
	\caption{\textbf{The overall framework of BEVHeight.} First, image-view encoder extracts high-dimensional image features. Then, image features are fed to the HeightNet to generate height distribution and context features, these two are further combined as fused features through an outer product operation. The height-based 2d to 3D projector push the fused features into an wedge-shaped 3D volume features. See \cref{alg:algorithm} for more details. Voxel Pooling splattes the wedge-shape features into an unified Bird's-Eye-View features, which are fed into the detection head to produce the final predictions. 
% 	\Tao{ 1.supervision? 2.height-based 2d->3d; 3.a little empty; 4.a little repeat with Figure 2 top; } \KY{Figure too large.}
	}
\label{fig:framework}
\end{figure*}

\subsection{Problem Definition}
\label{sec:problem_definition}
In this work, we would like to detect a three-dimensional bounding box of given foreground objects of interest. Formally, we are given the image $I\in R^{H\times W\times 3}$ from the roadside cameras, whose extrinsic matrix $E\in R^{3\times 4}$ and intrinsic matrix $K\in R^{3\times 3}$ can be obtained via camera calibration. 
We seek to precisely detect the 3D bounding boxes of objects on the image. We denote all bounding boxes of this image as $B=\left\{B_1,B_2,…,B_n\right\}$, and the output of detector as $\hat{B}$. 
% represents the annotations for each image used in the training stage. The detection results are defined
% as $\hat{B}_{ego}=\left\{\hat{B}_1,\hat{B}_2,…,\hat{B}_n\right\}$. 
Each 3D bounding box $B_{i}$ can be formulated as a vector with 7 degrees of freedom:
\begin{equation}
    \hat{B}_{i} = \left(x, y, z, l, w, h, \theta \right)
    \label{con:eq1}
\end{equation}
where $\left(x,y,z \right)$ is the location of each 3D bounding box . $\left(l,w,h \right)$ denotes the cuboid's length, width and height respectively. $\theta$ represents the yaw angle of each instance with respect to one specific axis. Specifically, a camera-only 3D object detector $F_{Det}$ can be defined as follows:
\begin{equation}
    \hat{B}_{ego} = F_{Det}\left(I_{cam}\right)
    \label{con:eq2}
\end{equation}
As a common assumption in autonomous driving, we assume the camera pose parameters $E$ and $K$ are known after the initial installation. In the roadside perception domain, people usually rely on multiple cameras installed at different locations to enlarge the perception range. This naturally encourages adopting those multi-view perception methods though the feature maps are not aligned geologically. Note that, although there are certain roadside units are equipped with other sensors, we focus on camera-only settings in this work for generalization purpose. 
% \KY{try to link why we focus on bev methods.}
% Meanwhile, 
% We expect the 3D object detector to maintain detection accuracy when the camera's posture changes inevitably and is calibrated.
% \ky{}

\subsection{Comparing the depth and height}
% \subsection{Delving into the Height}
\label{sec:delving_into_the_height}
As discussed before, state-of-the-art BEV camera-only methods first project the features into the bird's eye view space to reduce the z-axis dimension, then let the network learn implicitly \cite{liu2022petr, liu2022petrv2, li2022bevformer} or explicitly \cite{huang2021bevdet, li2022bevdepth, li2022bevstereo} about the 3D location information. Motivated by previous approaches in RGB-D recognition, one naive approach is to leverage the per-pixel depth as a location encoding.
In \cref{fig:histogram-depth-height}~(a), current methods firstly use an encoder to transform the original image into 2D feature maps. After predicting the per-pixel depth of the feature map, each pixel feature can be lifted into 3D space and zipped in the BEV feature space by voxel pooling techniques. 

However, we discover that using depth may be sub-optimal under the face-forwarding camera settings in autonomous driving scenarios. 
Specifically, we leverage the LiDAR point clouds of the DAIR-V2X-I~\cite{yu2022dair} dataset, where we first project these points to the images, to plot the histogram of per-pixel depth in \cref{fig:histogram-depth-height}~(b). We can observe a large range from 0 to 200 meters. By contrast, we plot the histogram of the per-pixel height to the ground and clearly observe the height ranges from -1 to 2m respectively, 
% We hypothesize this is easier for the network to predict.
which is easier for the network to predict.
But in practice, the predicted height can't be employed directly to the pinhole camera model like depth. 
How to achieve the projection from 2D to 3D effectively through height has not been explored.
% We ingeniously designed a virtual coordinate system and a reference plane (refer \cref{alg:algorithm} for details), which help us achieve the projection from 2D to 3D effectively. 

% To illustrate the benefits of the distribution of height to over that of depth, we compare the depth and height from the ground distribution of annotated instances from DAIR-V2X-I\cite{yu2022dair} Dataset in two aspects: (1) Histogram Distribution. (2) the distribution under the extrinsic disturbance.

% \mypara{Histogram Distribution.} Based on the 3D annotations of DAIR-V2X-I dataset, we analyze the histogram distribution of their depth and the height from the ground. As shown in \cref{fig:three}, height and depth approximately follow a normal distribution. The variance of height distribution is less than one ten-thousandth of that of the depth, which indicates that height estimation is more straightforward than depth.

\mypara{Analysis when extrinsic parameter changes.}
In \cref{fig:five}~(a), we provide an visual example of extrinsic disturbance.
To show that predicting height is superior to depth, we plot the scatter graph to show the correlation between the object's row coordinates on the image and its depth and height. Each plot represent an instance. As shown in \cref{fig:five}~(b). we observe that objects with smaller depths have a smaller $v$ value. However, suppose the extrinsic parameter changes; we plot the same metric in blue and observe that these values are drastically different from the clean setting. In that case, i.e., there is only a small overlap between the clean and noisy settings. We believe this is why the depth-based methods perform poorly when external parameters change. On the contrary, as observed in \cref{fig:five}~(c), the distribution remains similar regardless of the external parameter changes, i.e. the overlap between orange and blue dots is large. This motivates us to consider using height instead of depth. However, unlike depth that can be directly lifted to the 3D space via camera model, directly predicting height will not work to recover the 3D coordinate. Later, we present a novel height-based projection module to address this issue.

% we visualize the correlation between the image row coordinates of the object and its corresponding depth and height in the form of a scatter diagram in \cref{fig:five}. As shown, after adjusting the roll and pitch angle, the depth distribution only slightly overlaps before, while the height distribution has a much larger intersection. This phenomenon indicates that the correlation between the object's position on the image and its height from the ground is less affected by the roll and pitch perturbations. Thus, the height-based BEV detection method is less susceptible to the camera's external parameter perturbations.

\subsection{BEVHeight}
\label{sec:BEVHeight}

\mypara {Overall Architecture.}
As shown in Fig.\ref{fig:framework},  our proposed BEVHeight framework consists of five main stages.
% Image-view Encoder, HeightNet, $2D\rightarrow 3D$ projector, Voxel Pooling, and 3D Detection Head. 
 The image-view encoder that is composed of a 2D backbone and an FPN module aims to extract the 2D high-dimensional multi-scale image features $F^{2d} \in R^{C_F\times \frac{H}{16} \times \frac{W}{16}}$ given an image $I\in R^{3\times H \times W}$ in roadside view, where $C_F$ denotes the channel number. $H$ and $W$ represent the input image's height and width, respectively.
 The HeightNet is responsible for predicting the bins-like distribution of height from the ground $H^{pred} \in R^{C_H \times \frac{H}{16} \times \frac{W}{16}}$ and the context features $F^{context}\in R^{C_c \times \frac{H}{16}\times \frac{W}{16}}$ based on the image fractures $F^{2d}$, where $C_H$ stands for the number of height bins, $C_c$ denotes the channels of the context features. The fused features $F^{fused}$ that combines image context and height distribution is generated using Eq.~\ref{con:eq3}.
 The height-based $2D\rightarrow 3D$ projector pushes the fused features $F^{fused}$ into the 3D wedge-shaped features $F^{wedge} \in R^{X \times Y \times Z \times C_{c}}$ based on the predicted bins-like height distribution $H^{pred}$. See Algorithm \cref{alg:algorithm} for more details.
 Voxel Pooling transforms the 3D wedge-shaped features into the BEV features $F^{bev}$ along the height direction. 
 3D detection head firstly encodes the BEV features with convolution layers, And then predicts the 3D bounding box consisting of location $\left(x, y, z\right)$, dimension$\left(l, w, h\right)$, and orientation $\theta$.

\begin{equation}
    \begin{aligned}
        F^{fused} = F^{context}\otimes H^{pred},\\
        F^{fused} \in R^{C_c \times C_H \times \frac{H}{16} \times \frac{W}{16}}
    \end{aligned}
    \label{con:eq3}
\end{equation}

\mypara {HeightNet.}
% Our HeightNet follows the general design in BEVDepth~\cite{li2022bevdepth},
Motivated by the DepthNet in BEVDepth~\cite{li2022bevdepth}, we leverage a Squeeze-and-Excitation layer to generate the context features  $F^{context}$ from the 2D image features $F^{2d}$. Concretely, we stack multiple residual blocks~\cite{he2016deep} to increase the representation power and then use a deformable convolution layer~\cite{XizhouZhu2018DeformableCV} to predict the per-pixel height. We denote this height module as $H^{pred}$. To facilitate the optimization process, we translate the regression task to use one-hot encoding, i.e. discretizing the height into various height bins. The output of this module is  $h\in R^{C_H\times 1\times 1}$. Moreover, previous depth discretization strategies~\cite{HuanFu2018DeepOR,YunleiTang2020Center3DCM} are generally fixed and thus not suitable for roadside height predictions. To this end, we present an
% However, these methods' discretization bins are fixed and can't be adaptively adjusted according to the concerned height ranges in roadside scenarios. In this way, we adopt an
dynamic discretization as follow:
\begin{equation}
    h_i =  \lfloor{N \times \sqrt[\alpha]{ \frac{h-h_{min}}{h_{max}-h_{min}}}}\rfloor,
    \label{con:eq4}
\end{equation}
where $h$ represents the continuous height value from the ground, $h_{min}$ and $h_{max}$ represent the start and end of the height range. $N$ is the number of height bins, and $h_i$ denotes the value of $i-th$ height bin. $H$ is the height of the roadside camera from the ground. $\alpha$ is the hype parameter to control the concentration of height bins. See the supplementary material for more details.
% \KY{But honestly, i do not think this paragraph make sense by showing the sampling. Consider show a failure case of previous strategies and ours. otherwise I would suggest to combine this with previous height net. }

% whe context branch that 
% consists of a SE-like layer are used to generate the context features $F^{context}$ from the 2D image features $F^{2d}$. 
% The height layers composed of 
% stacked multiple Residual Blocks~\cite{he2016deep} and a Deformable Conv ~\cite{XizhouZhu2018DeformableCV} are responsible for predicting the height distribution $H^{pred}$. Each pixel value on the height distribution map is represented as $h\in R^{C_H\times 1\times 1}$, this means the probability distribution belongs to the pixel's discrete height bins. 

\begin{algorithm}[h!t]
\caption{Height-based 2D to 3D projector}
\label{alg:algorithm}
\textbf{Parameters Definition}: \\
   $O$ and $X,Y,Z$: coordinate system, where $O_{virt.}$ has the same origin as $O_{cam}$ with Y-axis prependicular to the ground.  \\
%   $O_{cam}XYZ$:  camera coordinate system.\\
%   $O_{virt.}XYZ$: virtual coordinate system, its origin coincides with that of the camera coordinate system, and its Y-axis is perpendicular to the ground.\\
%   $O_{ego}XYZ$: ego coordinate system of the roadside unit.\\
%   ref plane: the reference plane parallel to the image plane, the distance from the image plane is 1.\\
    $T_{A}^{B}$: transformation matrix from coordinate A to B. \\
    % $T_{cam}^{virt.}$: the transformation matrix from the camera's coordinate system to the virtual coordinate system.\\
    % $T_{cam}^{ego}$: the transformation matrix from the virtual coordinate system to the ego coordinate system.\\
    $K$: the camera's intrinsic matrix. \\
    $H$: the distance from the origin of the virtual coordinate system to the ground.\\
    $h_i$: the height from the ground of i-th height bin.\\
    $P_{ref}^{B}$: the pixel $(u,v)$ projected from reference plane A in coordinate B \\
    % $P_{ref}^{cam}$: the pixel $(u, v)$ projection point on the reference plane in the camera coordinate.\\
    % $P_{ref}^{virt.}$: the pixel $(u, v)$ projection point on the reference plane in the virtual coordinate system.\\
    $P_i^{A}$: the pixel $(u, v)$ projection point on i-th height bin in the coordinate system A.\\
    % $P_i^{virt.}$: the pixel $(u, v)$ projection point on i-th height bin in the virtual coordinate system.\\
    % $P_i^{ego}$: the pixel $(u, v)$ projection point on i-th height bin in the ego coordinate system.\\
    % $B_{ref}^{virt.}$: the projection point of $P_{ref}^{virt.}$ on the Y-axis of the virtual coordinate system.\\
    % $C_{ref}^{virt.}$: the projection point of $P_i^{virt.}$ on the Y-axis of the virtual coordinate system.\\
\vspace{-0.2cm}
\hrule
\vspace{0.1cm}
\textbf{Input}: \\
   $F^{fused} = \left\{f_1^{fused}, ...,f_{\frac{H}{16} \times \frac{W}{16}}^{fused}\right\}$, $f_m^{fused} \in R^{C_H \times C_c}$\\
   $H$; $K$; $T_{cam}^{virt.}$; $T_{cam}^{ego}$\\
\textbf{Output}: \\
    $F_{wedge}$ is the 3D wedge-shaped volume features.\\
% \vspace{-0.2cm}
% \hline
% \vspace{0.1cm}
\textbf{Begin:}
\begin{algorithmic}[1] %[1] enables line numbers
\STATE $F_{wedge} = \left\{\right\}$
\FOR {$f_m^{fused}$ in $F^{fused}$}
 \STATE $u, v \gets m$
 \STATE $P_{ref}^{cam}=K^{-1} [u,v,1]^T$\\
 \STATE $P_{ref}^{virt.}= \left\{x_{ref}^{virt.},y_{ref}^{virt.},z_{ref}^{virt.}\right\} = T_{cam}^{virt.} P_{ref}^{cam}$
 \FOR {$i\gets 0$ to $C_H$}
    %  \STATE $P_i^{virt.}=\frac{H-h_i}{y_{ref}^{virt.}}P_{ref}^{cam}$\\
    \STATE $P_i^{virt.}=\frac{H-h_i}{y_{ref}^{virt.}}P_{ref}^{virt.}$\\
     \STATE $P_i^{ego}= T_{virt.}^{ego}P_i^{virt.}$\\
     \STATE $F_{wedge} \gets F_{wedge} \cup associate(P_i^{ego}, f_m^{fused}[i])$\\
  \ENDFOR
\ENDFOR
\STATE \textbf{return} $F_{wedge}$\\
\end{algorithmic}
\textbf{End}
\end{algorithm}

\mypara{Height-based 2D-3D projection module.} 
Unlike the ``lift'' step in previous depth-based methods, one cannot recover the 3D location with only height information. To this end, we design a novel 2D to 3D projection module to push the fused features $F^{fused} \in R^{C_H \times C_c \times \frac{H}{16} \times \frac{W}{16}}$ into the wedge-shaped volume feature $F^{wedge} \in R^{X\times Y\times Z\times C_{c}}$ in the ego coordinate system.
% The 2D to 3D projector is applied to push the fused features $F^{fused} \in R^{C_H \times C_c \times \frac{H}{16} \times \frac{W}{16}}$ into the wedge-shaped volume feature $F^{wedge} \in R^{X\times Y\times Z\times C_{c}}$ in the ego coordinate system. 
% \ky{Unlike the} ``lift" step in previous depth-based methods, the height-fused features  $F^{fused}$ need ingenious transformations. 
As illustrated in \cref{fig:framework} and \cref{alg:algorithm}, we design a virtual coordinate system, with the origin coinciding with that of the camera coordinate system and the Y-axis perpendicular to the ground, and a special reference plane parallel to the image plane with a fixed distance 1. 

For each point $p_{image} = (u, v)$ in the image plane, we first choose the associated point $p_{ref}$ in the reference plane $plane_{ref}$, whose depth is naturally 1, i.e, $d_{ref}=1$. Thus we can project $p_{ref}$ from the uvd space to the camera coordinate through the camera's intrinsic matrix:
\begin{equation}
P_{ref}^{cam}=K^{-1} d_{ref} [u,v,1]^T = K^{-1} [u,v,1]^T . 
\end{equation}
Further, it can be transformed to the virtual coordinate to get $P_{ref}^{virt.}$ with the transformation matrix $T_{cam}^{virt.}$: 
\begin{equation}
    P_{ref}^{virt.} = T_{cam}^{virt.} P_{ref}^{cam}.
\end{equation}
Now we can know the point $p_{ref}$ in our virtual coordinate is $P_{ref}^{virt.}$.
% $P_{ref}^{virt.} = (x_{ref}^{virt.}, y_{ref}^{virt.}, z_{ref}^{virt.})$. 
Suppose the $i-th$ value in height bins relative to the ground for point $p_{image}$ is $h_i$ and the height from the origin of the virtual coordinate system to the ground is $H$. Based on similar triangle theory, we can have the $i-th$ projected 3D point in height virtual coordinate for $p_{image}$:
\begin{equation}
P_{i}^{virt.}=\frac{H-h_i}{y_{ref}^{virt.}} P_{ref}^{virt.} .
\end{equation}
Finally, we transform the $P_{i}^{virt.}$ to the ego-car space:
\begin{equation}
P_{i}^{ego}=T_{virt.}^{ego} P_{i}^{virt.} .
\end{equation}

% Summarize the above process, we can finish the 2D to 3D projection through:
In summary, the contribution of our module is in two-fold: i) we design a virtual coordinate system that leverages the height from the HeightNet; ii) we adopt a reference plane to simplify the computation by setting a constant depth to 1.  We formulate the height-based 2D-3D projection as follow:
\begin{equation}
    P_{i}^{ego}=T_{virt.}^{ego} \frac{H-h_i}{y_{ref}^{virt.}} T_{cam}^{virt.} K^{-1} [u,v,1]^T.
\end{equation}

% The arrows in \cref{fig:2dto3d} and the loop in \cref{alg:algorithm} also illustrate this projection is well-designed and efficient.

% Rethinking this process, there are two key elements worth aftertaste again: first one is the virtual coordinate system, which allows the HeightNet to directly predict the height with small variance relative to the ground; Another one is the reference plane, which simplifies the whole processing and expression through constant depth of 1.
% The overall process of 2D to 3D projector is depicted in \cref{alg:algorithm}.

% \mypara{Height discretization.}
% \KY{Should we group this with height net?}
% The height discretization can be performed with uniform discretization (UD) with a fixed bin size, spacing-increasing discretization (SID)~\cite{HuanFu2018DeepOR} with increasing bin sizes in logspace, or linear-increasing discretization (LID) ~\cite{YunleiTang2020Center3DCM} with linearly increasing bin sizes.
% The above four height discretization techniques are visualized in Fig. \ref{fig:seven}.
% Height supervision comes from the object-level annotations $B_{ego}$ and the optional ground-truth $H^{gt}$ derived from point clouds. \tao{this sentence mean?} \KY{I think we should remove this.}

% exp
\section{Experiments}
We briefly introduce the experiment settings and two benchmark datasets in road-side perception domain. We then compare our proposed \name{} with state-of-the-art methods under clean and noisy camera settings. We ablate our methods in detail and discuss the limitations. 
% In this section, we first introduce our experimental setups. Then, we conduct comprehensive experiments to validate the effects of our \name{} on two large-scale roadside datasets, DAIR-V2X~\cite{yu2022dair} and Rope3D~\cite{ye2022rope3d}. Finally, we also give an analysis on the popular vehicle perception dataset, nuScenes~\cite{caesar2020nuscenes}.

\subsection{Datasets}
\mypara{DAIR-V2X.} Yu \etal~\cite{yu2022dair} introduces a large-scale, multi-modality dataset. As the original dataset contains images from vehicles and roadside units, this benchmark consists of three tracks to simulate different scenarios. Here, we focus on the DAIR-V2X-I, which only contains the images from mounted cameras to study roadside perception. Specifically, DAIR-V2X-I contains around ten thousand images, where 50\%, 20\% and 30\% images are split into train, validation and testing respectively. However, up to now, the testing examples are not yet published, we evaluate the results on the validation set. We follow the benchmark to use the average perception of the bounding box as in KITTI~\cite{geiger2012we}.

% \mypara{DAIR-V2X~\cite{yu2022dair}} is a large-scale, multi-modality vehicle-infrastructure collaborative dataset for 3D object detection, which is composed of three sub-datasets: DAIR-V2X-C, DAIR-V2X-I, DAIR-V2X-V. DAIR-V2X-C containing sensors from both vehicle and roadside is for the vehicle-infrastructure cooperative 3D object detection. DAIR-V2X-V including sensors on the vehicle is for the 3D object detection in autonomous driving. DAIR-V2X-I with only sensors on infrastructure, is designed for the roadside perception task. In this paper, we conduct experiments based on the DAIR-V2X-I. DAIR-V2X-I contains 10084 samples, which are further divided into train/val/test according to 5:2:3 respectively. Since the test set is not released yet, all experiments are trained on the train set and evaluated on val set. The evaluation metric is average precision (AP) as used in the KITTI\cite{geiger2012we} dataset.

% Rope3D
\mypara{Rope3D\cite{ye2022rope3d}.} 
There is another recent large-scale benchmark named Rope3D. It contains over 500k images with three-dimensional bounding boxes from seventeen intersections. Here, we follow the proposed homologous setting to use 70\% of the images as training, and the remaining as testing. Note that, all images are randomly sampled. For validation metrics, we leverage the AP$_{\text{3D}{|\text{R40}}}$~\cite{simonelli2019disentangling} and the Rope$_\text{score}$, which is a consolidated metric of the 3D AP and other similarities metrics, such as average area similarity.

% is a large-scale dataset dedicated to vision-based roadside 3D object detection. This dataset provides 50009 frames of images together with 3D annotations. All images are sampled from seventeen intersections and split into the train/val set according to 7:3 under the Homologous setting. The proposed AP$_{\text{3D}{|\text{R40}}}$ and Rope$_\text{score}$ are used as the metrics.

% \mypara{nuScenes\cite{HolgerCaesar2019nuScenesAM}} is a large-scale autonomous-driving dataset for 3D detection, consisting of 700, 150 and 150 scenes for training, validation, and testing, respectively. 
% Each frame contains one point cloud and six calibrated images that cover 360 fields-of-view. 
% metric
% For 3D detection, the main metrics are mean Average Precision (mAP) and nuScenes detection score (NDS). 
% The mAP is defined by the BEV center distance with thresholds of {0.5m, 1m, 2m, 4m}, instead of the IoUs of bounding boxes. 
% NDS is a consolidated metric of mAP and other metric scores, such as average translation error and average scale error.

\subsection{Experimental Settings}
For architecture details, we use ResNet-101\cite{he2016deep} as image-view encoder in results compared with state-of-the-art and ResNet-50 for other ablation studies.
The input resolution is in (864, 1536). For data augmentation, we follow \cite{li2022bevdepth} to use random scaling and rotation in the 2D space only. All methods are trained for 150 epochs with AdamW optimzer~\cite{loshchilov2017adamw}, where the initial learning rate is set to $2e-4$.

\begin{table}[t]
 \scriptsize\centering\addtolength{\tabcolsep}{-4.2pt}
\caption{\textbf{Comparing with the state-of-the-art on the DAIR-V2X-I val set.} Here, we report the results of three types of objects, vehicle~(veh.), pedestrian~(ped.) and cyclist~(cyc.). Each object is categorized into three settings according to the difficulty defined in ~\cite{yu2022dair}. First, recent BEVDepth surpasses the previous best by a large margin, showing that using bird's-eye-view indeed helps in roadside scenarios. Our method outperforms the BEVDepth by over 3\% in average precision and constitutes state-of-the-art. It is surprising to see that our method outperforms those relying on LiDAR modality.
}
% \vspace{-0.2cm}

 \begin{tabularx}{1.0\linewidth}{l|c|ccc|ccc|ccc}
  \toprule
 \multirow{3}{*}{Method} &  
 \multirow{3}{*}{M}  
 & \multicolumn{3}{c|}{$\text{Veh.}_{(IoU=0.5)}$} & \multicolumn{3}{c|}{$\text{Ped.}_{(IoU=0.25)}$} & \multicolumn{3}{c}{$\text{Cyc.}_{(IoU=0.25)}$} \\
    \cmidrule(r){3-11}
     &  & Easy & Mid & Hard & Easy & Mid & Hard & Easy & Mid & Hard  \\
\midrule
PointPillars~\cite{lang2019pointpillars} & L &63.07 & 54.00 & 54.01 & 38.53 & 37.20 & 37.28 & 38.46 & 22.60 & 22.49 \\
SECOND~\cite{yan2018second} & L &71.47 & 53.99 & 54.00 & 55.16 & 52.49 & 52.52 & 54.68 & 31.05 & 31.19 \\
MVXNet~\cite{Sindagi2019MVX} & LC &71.04 & 53.71 & 53.76 & 55.83 & 54.45 & 54.40 & 54.05 & 30.79 & 31.06 \\
\midrule
ImvoxelNet~\cite{rukhovich2022imvoxelnet} &C & 44.78 & 37.58 & 37.55 & 6.81 & 6.746 & 6.73 & 21.06 & 13.57 & 13.17 \\
BEVFormer~\cite{li2022bevformer} & C 	&	61.37&	50.73&	50.73&	16.89&	15.82&	15.95	&22.16&	22.13&	22.06\\
BEVDepth~\cite{li2022bevdepth}&	C 	&	
75.50&	63.58&	63.67&	34.95&	33.42&	33.27& 55.67&	55.47&	55.34\\
\midrule
 \rowcolor{cyan!30} BEVHeight & C &	
 \textbf{77.78}&	\textbf{65.77}&	\textbf{65.85}&	\textbf{41.22}&	\textbf{39.29}&	\textbf{39.46}	&\textbf{60.23}&	\textbf{60.08}&	\textbf{60.54}\\
    \bottomrule
\multicolumn{8}{l}{\scriptsize{M, L, C denotes modality, LiDAR, camera respectively.}}
  \end{tabularx}
  \vspace{-0.5cm}
  \label{dair_sota_2}
\end{table}

\subsection{Comparing with state-of-the-art}
\mypara{Results on the original benchmark.} On DAIR-V2X-I setting, we compare our BEVHeight with other state-of-the-art methods like ImvoxelNet~\cite{rukhovich2022imvoxelnet}, BEVFormer~\cite{li2022bevformer}, BEVDepth~\cite{li2022bevdepth} on DAIR-V2X-I val set. Some results of LiDAR-based and multimodal methods reproduced by the original DAIR-V2X~\cite{yu2022dair} benchmark are also displayed.
As can be seen from Tab.~\ref{dair_sota_2}, the proposed \name{} surpasses state-of-the-art methods by a significant margin of 2.19\%, 5.87\% and 4.61\% in vehicle, pedestrian and cyclist categories respectively.

% \mypara{Rope3D val set.}
On Rope3D dataset, we also compare our BEVHeight with other leading BEV methods, such as BEVFormer~\cite{li2022bevformer} and BEVDepth~\cite{li2022bevdepth}. Some results of the monocular 3D object detectors are revised by adapting the ground plane. As shown in Tab.~\ref{tab_performance_overall},  
we can see that our method outperforms all BEV and monocular methods listed in the table. In addition, under the same configuration, our BEVHeight outperforms the BEVDepth by $4.97\%$ / $4.02\%$, $3.91\%$ / $3.06\%$ on AP$_{\text{3D}{|\text{R40}}}$ and Rope$_\text{score}$  for car and big vehicle respectively.

\begin{table}[t]
\footnotesize
  \centering\addtolength{\tabcolsep}{-3.8pt}
\caption{\textbf{Results on the Rope3D val set.} Here, we follow~\cite{ye2022rope3d} to report the results on vehicles. Our method on average surpasses the state-of-the-art method over a margin of 3\% in both average precision and $Rope_{score}$ metric.
% \Tao{more caption}
}
\vspace{-0.2cm}
 \begin{tabularx}{1.\linewidth}{ l |cc|cc|cc|cc }
\toprule
\multirow{4}{*}{Method}   & \multicolumn{4}{c|}{IoU = 0.5} & \multicolumn{4}{c}{IoU = 0.7} \\ 
\cmidrule(r){2-9}
  & \multicolumn{2}{c|}{Car} & \multicolumn{2}{c|}{Big Vehicle} & \multicolumn{2}{c|}{Car} & \multicolumn{2}{c}{Big Vehicle} \\ 
\cmidrule(r){2-9}
&AP & Rope &
AP & Rope &
AP & Rope &
AP & Rope \\
% &AP$_{\text{3D}{|\text{R40}}}$ & Rope$_\text{score}$ & AP$_{\text{3D}{|\text{R40}}}$ & Rope$_\text{score}$  & AP$_{\text{3D}{|\text{R40}}}$ & Rope$_\text{score}$ & AP$_{\text{3D}{|\text{R40}}}$ & Rope$_\text{score}$\\ 

\midrule

M3D-RPN~\cite{brazil2019m3d} 
&54.19 & 62.65	&33.05 &  44.94 &16.75 & 32.90 &6.86  &  24.19 \\

Kinematic3D~\cite{brazil2020kinematic}  &50.57  & 58.86&	37.60&  48.08 &17.74  & 32.9 &   6.10&   22.88\\

MonoDLE~\cite{ma2021delving} 
 & 51.70 & 60.36 & 40.34 & 50.07  & 13.58 & 29.46 &9.63 &25.80\\

MonoFlex~\cite{zhang2021objects} & 60.33 & 66.86&	37.33 &47.96   & 33.78 & 46.12 &  10.08 &26.16\\

BEVFormer~\cite{li2022bevformer}	&50.62&	58.78&	34.58&	45.16&	24.64&	38.71	&10.05&	25.56\\

BEVDepth~\cite{li2022bevdepth}	&69.63&	74.70&	45.02&	54.64&	42.56&	53.05	&21.47	&35.82\\

\midrule
 \rowcolor{cyan!30} BEVHeight & \textbf{74.60}& \textbf{78.72}& \textbf{48.93}& \textbf{57.70}& \textbf{45.73}& \textbf{55.62}& \textbf{23.07}& \textbf{37.04} \\							
\bottomrule
\multicolumn{9}{l}{\footnotesize{AP and Rope denote AP$_{\text{3D}{|\text{R40}}}$ and Rope$_\text{score}$ respectively.}}
\end{tabularx}
% }
% \caption{Overall performance of the monocular 3D object detection approaches on the Rope3D Dataset with IoU = 0.5 and 0.7. $(G)$ denotes adapting the ground plane.}
 \vspace{-0.50cm}
\label{tab_performance_overall}
\end{table}

% 地面方程咋用的

\mypara{Results on noisy extrinsic parameters.}
% To verify the robustness of our BEVHeight when the camera's extrinsic matrix is changed inevitably. 
In the realistic world, camera parameters frequently change for various reasons. Here we evaluate the performance of our framework in such noisy settings. We follow \cite{yu2022benchmarking} to simulate the scenarios that external parameters are changed. Specifically, we introduce a random rotational offset in normal distribution $N(0, 1.67)$ along the roll and pitch directions as the mounting points usually remain unchanged.  
% \KY{degree of what?}

% This situation often occurs during the maintenance of roadside cameras. In this case, the camera's extrinsic matrix will differ from its previous state when the labeled data is collected. The generalization to the camera's mount position disturbance is a great challenge for the existing methods.
During the evaluation, we add the rotational offset along roll and pitch directions to the original extrinsic matrix. The image is then applied with rotation and translation operations to ensure the calibration relationship between the new external reference and the new image. Examples are given in Sec.~\ref{sec:visualization_results}.
As shown in Tab.~\ref{dair_robust}, the performance of the existing methods degrades significantly when the camera's extrinsic matrix is changed. Take $\text{Vehicle}_{(IoU=0.5)}$ for example, the accuracy of BEVFormer~\cite{li2022bevformer} drops from 50.73\% to 16.35\%. The decline of BEVDepth~\cite{li2022bevdepth} is from 60.75\% to 9.48\%, which is pronounced. Compared with the above methods, Our BEVHeight maintains 51.77\% from the original 63.49\%, which surprises the BEVDepth by 42.29\% on vehicle category.

\mypara{Visualization Results.}
\begin{figure*}[t!]
	\centering
	\includegraphics[width=\textwidth]{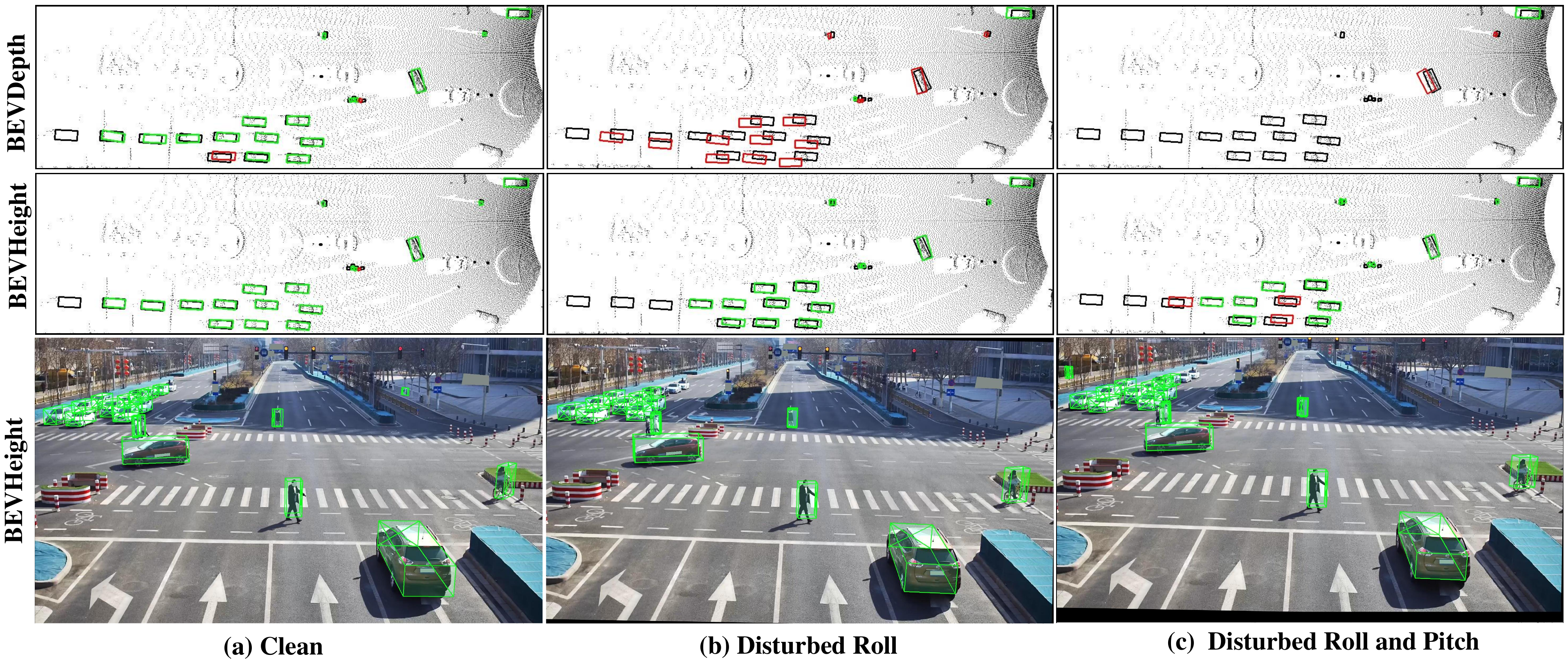}
	\caption{\textbf{Visualization Results of BEVDepth and our proposed BEVHeight under the extrinsic disturbance.} We use boxes in \textbf{{\color{red}red}} to represent false positives,  \textbf{{\color{green}green}} boxes for truth positives, and \textbf{{\color{black}black}} for the ground truth. The truth positives are defined as the predictions with IoU\textgreater 0.5 for vehicle and IoU\textgreater 0.25 for pedestrian and cyclist. (a) Clean means the original image without any processing; (b) Disturbed Roll denotes camera rotate 1 degree along roll direction; (c) Disturbed Roll and Pitch represents camera rotate 1 degree along roll and pitch directions simultaneously. 
% 	The image is processed with rotation and translation operations to maintain an accurate calibration relationship between the new image and the new camera's external matrix.
We observe that our methods outperform the baseline in all three settings. Note that, the BEVDepth only identify two objects under roll-pitch disturbance while ours identify nine. }

\label{fig:visualization}
% \vspace{-0.2cm}
\end{figure*}
\label{sec:visualization_results}
As shown in Fig.~\ref{fig:visualization}, we present the results of BEVDepth~\cite{li2022bevdepth} and our BEVHeight in the image view and BEV space, respectively. The above two models are not applied with data augmentations in the training phase. From the samples in (a), we can see that the predictions of BEVHeight fit more closely to the ground truth than that of BEVDepth. As for the results in (b), under the disturbance of roll angle, there is a remarkable offset to the far side relative to the ground truth in BEVDepth detections. In contrast, the results of our method are still keeping the correct position with ground truth.  Moreover, referring to the predictions in (c), BEVDepth can hardly identify far objects, but our method can still detect the instance in the middle and long-distance ranges and maintain a high IoU with the ground truth.

\begin{table}[t]
 \scriptsize\centering\addtolength{\tabcolsep}{-3.25pt}
   \caption{
%   \textbf{The robustness under extrinsic parameter perturbations on DAIR-V2X-I.} \Tao{more caption}
    \textbf{Results on robustness settings. } Here, we simulate the robustness scenarios where the external parameters of the camera changes. Consider the  Specifically, we consider two degrees of freedom mutation, roll and pitch of the camera center. In both dimensions, we randomly sample angles from a normal distribution of $\mathcal{N}(0, 1.67)$. Surprisingly, given such minor changes, traditional depth-based methods decrease to under 15\% even for those vehicles under easy settings. On the contrary, our methods achieve around 577\% improvement compared to those baselines, evidencing the robustness of \name{}.
%   “roll” and “pitch” means applying an additional rotation offset in normal distribution N(0, 1.67) to the camera’s extrinsic matrix along roll and pitch directions
   }
  %\vspace{-0.1cm}
 \begin{tabularx}{1.0\linewidth}{l|cc|ccc|ccc|ccc}
  \toprule
 \multirow{3}{*}{\rotatebox{90}{Model}} & \multicolumn{2}{c|}{Disturbed}
    \ & \multicolumn{3}{c|}{$\text{Veh.}_{(IoU=0.5)}$} & \multicolumn{3}{c|}{$\text{Ped.}_{(IoU=0.25)}$} & \multicolumn{3}{c}{$\text{Cyc.}_{(IoU=0.25)}$}  \\
   \cmidrule(r){2-12}
   & roll &	pitch & Easy & Mid & Hard & Easy & Mid & Hard & Easy & Mid & Hard  \\
  
    \midrule
 \multirow{4}{*}{\rotatebox{90}{BEVFormer}} &&&	61.37&	50.73	&50.73&	16.89&	15.82&	15.95	&22.16&	22.13&	22.0\\
	&	\checkmark& 	&	50.65&	42.9&	42.95&	10.16&	9.41&	9.47&	13.62&	13.71&	13.08\\
	&		&\checkmark&	46.40&	38.26&	38.37&	9.12	&8.44&	8.55&	8.99 &	8.43&	8.42 \\
 	&	\checkmark&	\checkmark&	19.24&	16.35&	16.47&	3.93&	3.43&	3.52&	4.93&	4.98&	4.98\\
\midrule
 \multirow{4}{*}{\rotatebox{90}{BEVDepth}}	& & &			71.56& 	60.75&	60.85&	21.55&	20.51&	20.75&	40.83	&40.66&	40.26\\
&	\checkmark	&&	34.82&	28.32&	28.35&	4.49&	4.36&	4.39&	10.48&	9.51&	9.73\\
	&&	\checkmark&	14.04&	11.41&	11.49&	3.01&	2.67&	2.75&	6.43&	6.23&	6.83\\
	&	\checkmark &\checkmark &	11.84&	9.48&	9.54&	2.16&	1.84&	1.89&	4.31&	4.14&	4.26\\
\midrule
 \multirow{4}{*}{\rotatebox{90}{BEVHeight}}	&	&&	75.58&	63.49&	63.59&	26.93&	25.47&	25.78&	47.97	& 47.45	& 48.12	\\
	&	\checkmark &&	66.06&	54.99&	55.14&	18.66&	17.63&	17.78&	34.45&	26.93&	27.68\\
	&&	\checkmark&	68.49&	56.98&	57.11&	17.94&	16.87&	17.09&	34.48&	27.82&	28.67\\
	&	\checkmark &	\checkmark&	62.64&	51.77&	51.9&	14.38&	14.01&	14.09&	31.28&	25.24&	26.02\\

    \bottomrule
    % \multicolumn{21}{l}{Mid: Middle, Veh.: Vehicle, Ped.: Pedestrian, Cyc.: Cyclist.}
  \end{tabularx}

  \label{dair_robust}
\end{table}

\begin{table}[t]
 \scriptsize\centering\addtolength{\tabcolsep}{-3.2pt}
 \caption{\textbf{Ablating our dynamic discretization on DAIR-V2X-I dataset.} Compared to the uniform discretization(UD), our method achieves on average 1\% improvement in average precision. 
%  \KY{add metric name here...}
%  \textbf{Ablation study on the effectiveness of PID in BEVHeight on DAIR-V2X-I.}\Tao{more caption}
 % \vspace{-0.1cm}
 }
 \begin{tabularx}{1.0\linewidth}{cc|ccc|ccc|ccc}
  \toprule
   \multicolumn{2}{c|}{Spacing}
%  \multicolumn{2}{c|}{\multirow{2.8}{*}{Spacing}} 
%  & \multicolumn{9}{c}{BEVHeight} \\
%     \cmidrule(r){3-11}
%   & 
   & \multicolumn{3}{c|}{$\text{Veh.}_{(IoU=0.5)}$} & \multicolumn{3}{c|}{$\text{Ped.}_{(IoU=0.25)}$} & \multicolumn{3}{c}{$\text{Cyc.}_{(IoU=0.25)}$} 
\\
   \cmidrule(r){1-11}
    DID ($\alpha$) &	UD & Easy & Mid & Hard & Easy & Mid & Hard & Easy & Mid & Hard  
   \\
    \midrule
   	~ & \checkmark &	75.63&	63.75&	63.85&	25.82&	25.47&	25.35&	47.52&	47.47&	47.19\\
   	\checkmark (1.5) &	~&	76.24&	64.54&	64.13&	26.47&	25.79&	25.72&	48.55&	48.21&	47.96\\
     \checkmark (2.0)& ~&	\textbf{76.61}&	\textbf{64.71}&	\textbf{64.76}&	\textbf{27.34}&	\textbf{26.09}&	\textbf{25.33}&	\textbf{49.68}&	\textbf{48.84} & \textbf{48.58}\\
    \bottomrule
  \end{tabularx}
  \vspace{-0.3cm}
  \label{dair_discretization}
\end{table}

\subsection{Ablation Study}
% \mypara{Robust to Extrinsic Disturbance.}

\mypara{Dynamic Discretization.}
Experiments in Tab.~\ref{dair_discretization} show the detection accuracy improvement 0.3\% - s1.5.0\% when our dynamic discritization is applied instead of uniform discretization(UD).
The performance when hype-parameter $\alpha$ is set to 2.0 suppresses that of 1.5 in most cases, which signifies that hype-parameter $\alpha$ is necessary to achieve the most appropriate discretization.

\mypara{Analysis on Point Cloud Supervision.}
\begin{table}[t]
 \scriptsize\centering\addtolength{\tabcolsep}{-2.9pt}
 \caption{\textbf{Results with point cloud supervision on DAIR-V2X-I dataset.} We can observe that for both BEVDepth and BEVHeight, LiDAR point cloud supervision did not help in terms of evaluation results. This is another evidence that road-side perception is different from the ego-vehicle one.  }
% \vspace{-0.2cm}
 \begin{tabularx}{1.0\linewidth}{l|ccc|ccc|ccc}
  \toprule
 \multirow{2}{*}{Method}    & \multicolumn{3}{c|}{$\text{Veh.}_{(IoU=0.5)}$} & \multicolumn{3}{c|}{$\text{Ped.}_{(IoU=0.25)}$} & \multicolumn{3}{c}{$\text{Cyc.}_{(IoU=0.25)}$}  \\
   \cmidrule(r){2-10}
 & Easy & Mid & Hard & Easy & Mid & Hard & Easy & Mid & Hard  
 \\
    \midrule
    BEVDepth	& 71.56& 	60.75&	60.85&	21.55&	20.51&	20.75&	40.83	&40.66&	40.26\\
    BEVDepth$\dagger$	&	71.09&	60.37&	60.46&	21.23&	20.84&	20.85&	40.54&	40.34&	40.32\\
    \midrule
    BEVHeight	 &	75.58&	63.49&	63.59&	26.93&	25.47&	\textbf{25.78}&	47.97	& 47.45	& 48.12	\\
    BEVHeight$\dagger$	& \textbf{75.64}& \textbf{63.61}&	\textbf{63.72}&	\textbf{27.01}&	\textbf{25.55}&	25.34&	\textbf{48.03}&	\textbf{47.62}&	\textbf{48.19}\\
    \bottomrule
    \multicolumn{10}{l}{\scriptsize{$\dagger$ denotes training with PointCloud supervision.}}
  \end{tabularx}
  \label{pc_sup}
\vspace{-0.55cm}
\end{table}

To verify the effectiveness of point cloud supervision in roadside scenes, we conduct ablation experiments on both BEVDepth~\cite{li2022bevdepth} and our method. As shown in Tab.~\ref{pc_sup}, BEVDepth with point cloud supervision is slightly lower than that without supervision. As for our BEVHeight, although there is a slight improvement under the IoU=0.5 condition, the overall gain is not apparent. This can be explained by the fact that the background in roadside scenarios is stable. These background point clouds fail to provide adequate supervised information and increase the difficulty of model fitting.
% there is only a slight improvement under IoU=0.5. We speculate this is because the camera is fixed in roadside scenarios, thus the majority of the pixels belong to the background and have relatively fixed depth or height values. The network can learn them well even without ground truth as supervision.
% We speculate that this is due to the fact that the background in roadside scenarios is stable. 
% These background point clouds remain unchanged and fail to provide adequate supervised information
% and increase the difficulty of model fitting.
% 分析，depth的问题，猜测不同路口混合训练的影响，加了监督更加hard让网络拟合了

% \mypara{Analysis on point cloud supervision.}

\begin{figure}[t!]
	\centering
	\includegraphics[width=8.5cm]{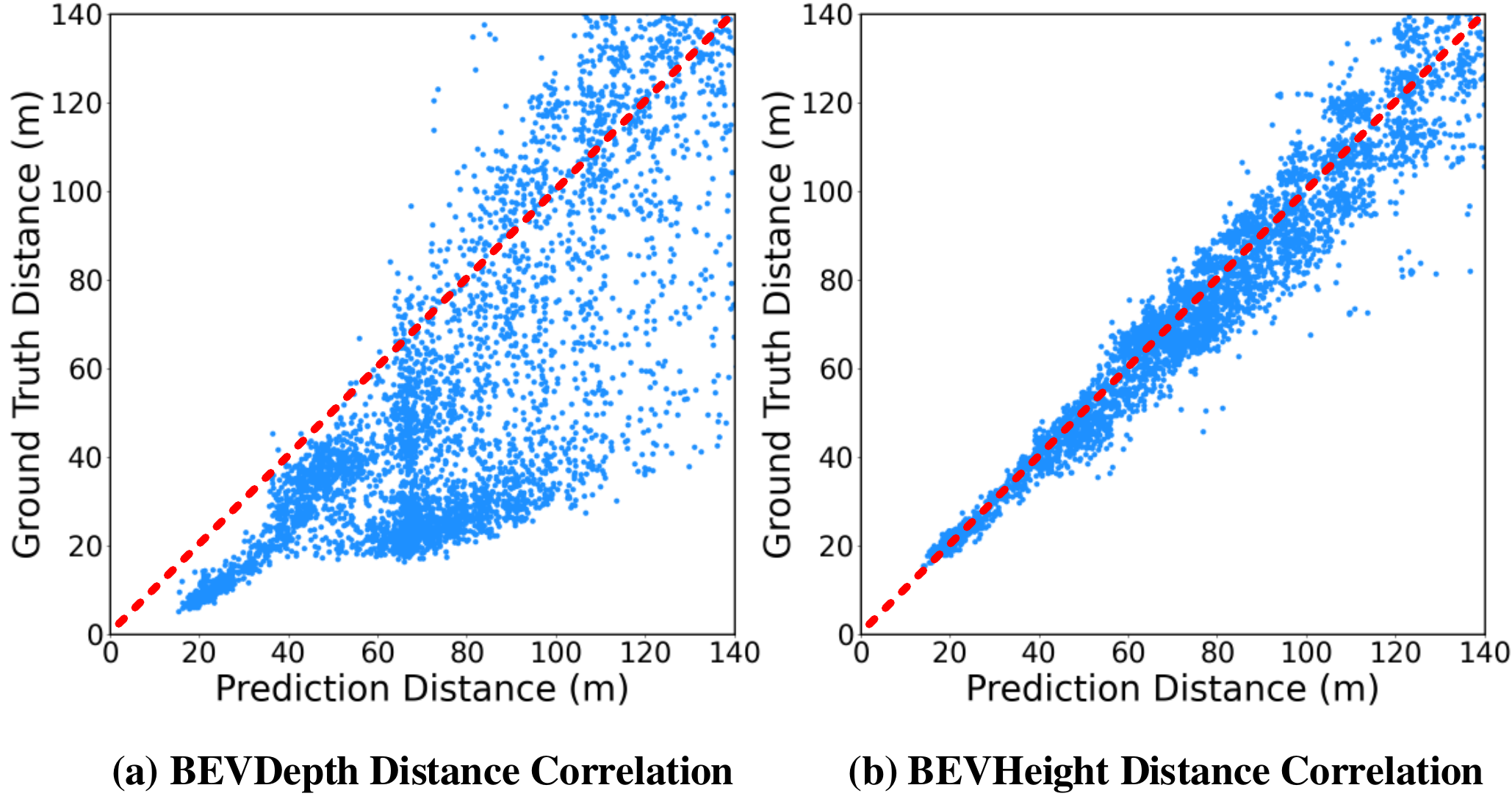}
	\caption{\textbf{Empirical analysis of the distance correlation.} All experiments are conducted on the DAIR-V2X-I val set. (a) and (b) reveal the distance correlation between ground truth and predicted distance on the BEVDepth and our BEVHeight. We take distances from the camera's coordinate system origin to the annotated objects' center for consideration. Each point represents an annotated instance. The scatter diagram of BEVHeight in (b) is closer to the diagonal than that of BEVDepth in (a), indicating that the distance error triggered by height estimation is more minimal than the depth candidate.}

\label{fig:distance_correlation}
\vspace{-0.5cm}
\end{figure}

\mypara{Analysis on Distance Error.}
To provide a qualitative analysis of depth and height estimations, we convert depth and height to the distance between the predicted object's center and the camera’s coordinate origin, as is shown in Fig.~\ref{fig:distance_correlation}.  Compared with the distance error triggered by depth estimation in BEVDepth\cite{li2022bevdepth}, the height estimation in our BEVHeight introduces less error, which illustrates the superiority of height estimation over the depth estimation in the roadside scenario.

\mypara{Latency.}
As shown in Tab.~\ref{latency_rebuttal}, we benchmark the runtime of BEVHeight and BEVDepth. With an image size of 864x1536, BEVDepth runs at 14.7 FPS with a latency of 68ms, while ours runs at 16.1 FPS with 62ms, which is around 5\% faster. It is due to the depth range (1$\sim$104m) being much larger than height (-1$\sim$1m), thus ours has 90 height bins that less than 206 depth ones,
leading to a slimmer regression head and fewer pseudo  points for voxel pooling. It evidences the superiority of predicting height instead of depth and advocates the efficiency of our method.
\begin{table}[h!t]
\scriptsize\centering\addtolength{\tabcolsep}{-2.0pt}
\renewcommand\arraystretch{1.0}
\caption{{\bf Latency of BEVHeight and BEVDepth.} }
\begin{tabular}{l|c|c|c|c|c}
\toprule   
Methods& Backbone &Range & Number of bins & Latency (ms) & FPS \\ 
\midrule
BEVDepth~\cite{li2022bevdepth} & R50 & 1 - 104m& 206& 82& 12.2\\
\rowcolor{cyan!30} BEVHeight& R50 & -1 - 1m&  90& 77& 13.0\\
\midrule
    BEVDepth~\cite{li2022bevdepth} &  R101& 1 - 104m& 206& 68& 14.7\\
\rowcolor{cyan!30}	BEVHeight  & R101& -1 - 1m&  90& 62& 16.1\\
\bottomrule 
\multicolumn{6}{l}{\scriptsize Measured on a V100 GPU. Image shape 864×1536.}
\end{tabular}
\label{latency_rebuttal}
\vspace{-0.25cm}
\end{table}

\mypara{Limitations and Analysis.}
Though the motivation of our work is to address the challenges in the roadside scenarios, we nonetheless benchmark our methods on nuScenes to study the effectiveness. Here, the input resolution is set to (256, 704). We follow the setting of BEVDepth, i.e. the training lasts for 24 epochs. Note that, we did not use other tricks such as class-balanced grouping and sampling~(CBGS) strategy~\cite{zhu2019cbgs}, exponential moving average or multi-frame fusion. 
In \cref{tab:nus}, we observe that our method falls behind the BEVDepth by around 0.02 in mAP metrics. This shows that our method has limited performance on ego-vehicle settings. 

% \Tao{256x704 128x128 pc sup}
% \KY{to finish}
% For the nuScenes\cite{HolgerCaesar2019nuScenesAM} dataset, the input image is scaled to 256x704; we adopt the same dataset augmentations on image-view and BEV features as in BEVDepth\cite{li2022bevdepth}. All experiments are trained for 24 epochs without using the CBGS strategy, EMA, and multi-frame fusion.

\begin{table}[!t]
 \centering\addtolength{\tabcolsep}{-4.15pt}
\footnotesize
\caption{\textbf{Limitation of our method.} We present the results on the nuScenes validation dataset. We notice that our methods fall behind the traditional BEVDepth on the ego-vehicle settings by 2\%. This shows that our methods are effective on cameras with high installation and bird's-eye-view as in the roadside scenario, and is not ideal on cameras mounted on ego-vehicles.}
% \vspace{-0.1cm}
\begin{tabularx}{1.0\linewidth}{l|cccccccccc}
\toprule
 Method  &
			 mAP$\uparrow$ & NDS$\uparrow$ & mATE$\downarrow$ & mASE$\downarrow$  & mAOE$\downarrow$ & mAVE$\downarrow$ & mAAE$\downarrow$   \\
\midrule
\textcolor{gray!80}{BEVDepth}	&\textcolor{gray!80}{	0.315}&	\textcolor{gray!80}{0.367}&	\textcolor{gray!80}{0.702}&\textcolor{gray!80}{	0.271}&	\textcolor{gray!80}{0.621}&	\textcolor{gray!80}{1.042}&\textcolor{gray!80}{	0.315}\\
BEVDepth*	&		0.313&	0.354&	0.713&	0.280&	0.655&	1.230	&0.377\\
\midrule
BEVHeight	&	0.291&	0.342&	0.722&	0.278&	0.674&	1.230&	0.361\\
\bottomrule
\multicolumn{10}{l}{\footnotesize{* denotes the results we reproduce.}}
\end{tabularx}
\label{tab:nus}
\vspace{-0.60cm}
\end{table}

\begin{figure}[!h]
\centering
% \vspace{-0.2cm}
\includegraphics[width=8.5cm]{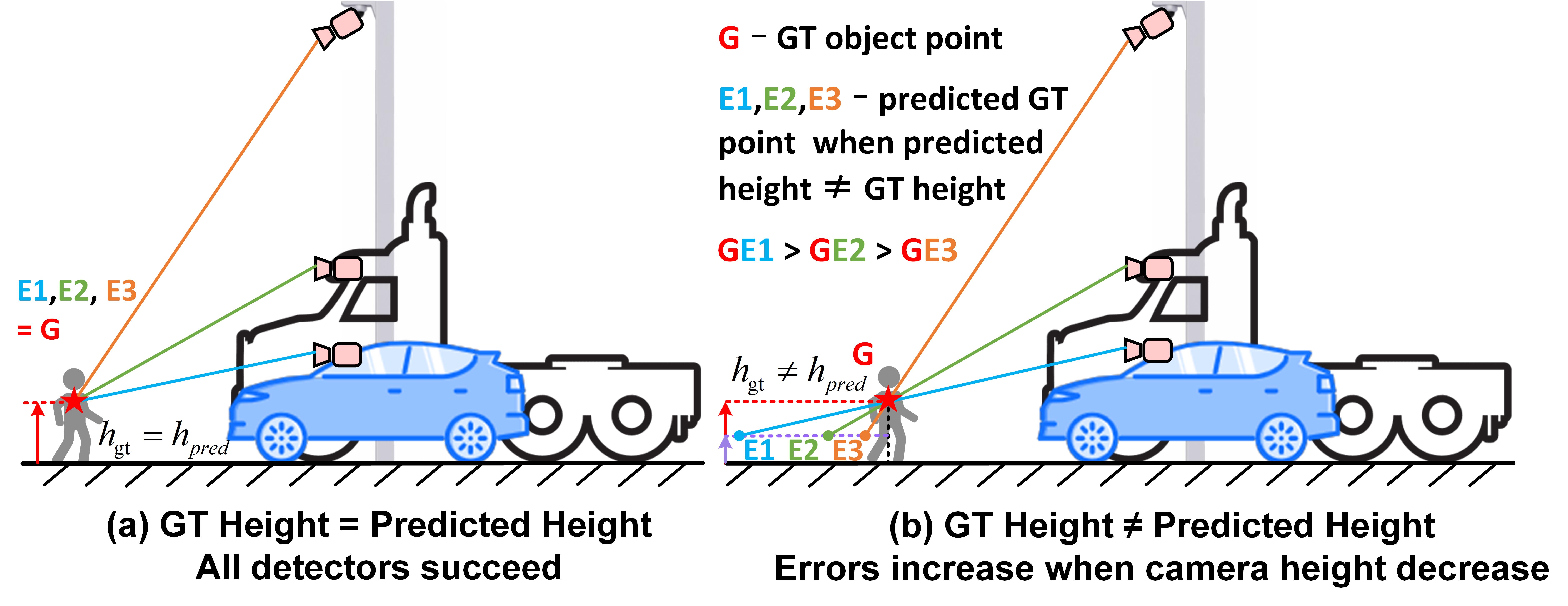}
\vspace{-0.60cm}
\caption{\textbf{Distance error analysis caused by same height estimation error on different platform cameras.}}
\vspace{-0.50cm}
\label{fig:versatility_analysis}
\end{figure}

Firstly, our method does \emph{not} assume the ground-plane is fixed, and it is not the reason why our method cannot surpass the depth-based one on ego-vehicle settings. To verify, we collect around 13 thousand sequences from the camera mounted on a moving truck with a ground height of 3.14m, and annotate the 3D object box following nuScenes. As shown in Tab.~\ref{ddd_rebuttal} We observe that our BEVHeight again surpasses the depth-based state-of-the-art by a large margin, evidences the performance is affected by the camera height but not time-varying ground plane and it can work on ego-vehicle settings.
We visualize three cameras observing the same object and analyze the detection error in Fig.~\ref{fig:versatility_analysis}: (a) shows when the height prediction is equal to the ground-truth, detection is perfect for all cameras; (b) if not, for the same height prediction error, the distance between the predicted point and ground-truth is inversely proportional to the camera ground height. This is why BEVHeight achieves on-par performance on nuScenes but quickly surpasses BEVDepth~\cite{li2022bevdepth} when the camera height only increases less than 1 meter.
% \begin{table*}[ht]
%  \centering\addtolength{\tabcolsep}{-0.6pt}
%  \resizebox{0.8\textwidth}{!}{
%  \begin{tabularx}{1.0\textwidth}{l|c|ccc|ccc|ccc}
%   \toprule
%  \multirow{3}{*}{Method} &  
%  \multirow{3}{*}{Modality}  
%  & \multicolumn{3}{c|}{$\text{Vehicle}_{(IoU=0.5)}$} & \multicolumn{3}{c|}{$\text{Pedestrian}_{(IoU=0.25)}$} & \multicolumn{3}{c}{$\text{Cyclist}_{(IoU=0.25)}$} \\
%     \cmidrule(r){3-11}
%      &  & Easy & Mid & Hard & Easy & Mid & Hard & Easy & Mid & Hard  \\
% \midrule

% PointPillars~\cite{lang2019pointpillars} & PointCloud &63.07 & 54.00 & 54.01 & 38.53 & 37.20 & 37.28 & 38.46 & 22.60 & 22.49 \\
% SECOND~\cite{yan2018second} & PointCloud &71.47 & 53.99 & 54.00 & 55.16 & 52.49 & 52.52 & 54.68 & 31.05 & 31.19 \\
% MVXNet~\cite{Sindagi2019MVX} & Image+PointCloud &71.04 & 53.71 & 53.76 & 55.83 & 54.45 & 54.40 & 54.05 & 30.79 & 31.06 \\
% \midrule
% ImvoxelNet~\cite{rukhovich2022imvoxelnet} &Image & 44.78 & 37.58 & 37.55 & 6.81 & 6.746 & 6.73 & 21.06 & 13.57 & 13.17 \\
% BEVFormer-R101$\ast$~\cite{li2022bevformer} & Image 	&	61.37&	50.73&	50.73&	16.89&	15.82&	15.95	&22.16&	22.13&	22.06\\
% BEVDepth-R101$\ast$~\cite{li2022bevdepth}&	Image 	&	76.01&	64.11&	64.18&	24.32&	24.96&	24.84	&46.45&	45.56&	45.69	\\

% \midrule
% BEVHeight-R101(Ours) & Image &	79.12&	67.95&	67.04&	29.85&	29.31&	29.07	&51.55&	51.39&	50.91\\
%     \bottomrule
%   \end{tabularx}
%   }
%   \caption{\textbf{Comparison on the DAIR-V2X-I val set.}}
%   \label{dair_sota}
% \end{table*}

\begin{table}[h!t]
 \scriptsize\centering\addtolength{\tabcolsep}{1.0pt}
\caption{\textbf{Experiments on the dataset collected by higher truck.}} 
 \resizebox{1.0\linewidth}{!}{
 \begin{tabularx}{1.0\linewidth}{l|ccc|ccc}
 \toprule
 \multirow{3}{*}{Method} &
\multicolumn{3}{c|}{$\text{Car}_{(IoU=0.5)}$} & \multicolumn{3}{c}{$\text{Big Vehicle}_{(IoU=0.5)}$} \\
 \cmidrule(r){2-7}
   & Easy & Mod. & Hard & Easy & Mod. & Hard \\
 \midrule
 BEVDepth ~\cite{li2022bevdepth} &	50.05 &	 36.82 &	36.82&	30.15&	24.74&	24.74	\\
 \rowcolor{cyan!30}BEVHeight & \textbf{51.77}&	\textbf{40.96}&	\textbf{40.96}&	\textbf{34.65}&	\textbf{29.01}&	\textbf{29.01}\\
\bottomrule
\end{tabularx}
}
\label{ddd_rebuttal}
\end{table}

\mypara{Contributions.}
Theoretically, our proposed height-based pipeline entails: i) representation agnostic to distance, as visualized in Fig.~\ref{fig:teaser}, ii) friendly prediction owing to centralized distribution as displayed in Fig.~\ref{fig:histogram-depth-height}, iii) robustness against extrinsic disturbance as illustrated in Fig.~\ref{fig:five}. Technically, we design a novel HeightNet and the projection module with less computational cost. Experimentally, experiments on various datasets and multiple depth-based detectors show the superiority of our method in both accuracy and latency.

\section{Conclusion}
We notice that in the domain of roadside perception, the depth difference between the foreground object and background quickly shrinks as the distance to the camera increases, this makes state-of-the-art methods that predict depth to facilitate vision-based 3D detection tasks sub-optimal. On the contrary, we discover that the per-pixel height does not change regardless of distance. To this end, we propose a simple yet effective framework, \name{}, to firstly predict the height and then project the 2D feature to 3D space to improve the detector. Through extensive experiments, \name{} surpasses BEVDepth baseline by a margin of 4.85\% and 4.43\% on DAIR-V2X-I and Rope3D benchmarks under the traditional clean settings, and by 26.88\% on robust settings where external camera parameters changes. We hope our work can shed light on studying more effective feature representation on roadside perception.

\section*{Acknowledgments}
This work was supported by the National High Technology Research and Development Program of China under Grant No. 2018YFE0204300, the National Natural Science Foundation of China under Grant No. 62273198, U1964203, the China Postdoctoral Science Foundation (No. 2021M691780).

%%%%%%%%% REFERENCES
{\small
\normalem
\bibliographystyle{ieee_fullname}
\bibliography{egbib}
}

\clearpage
\begin{appendices}
\section{Appendix}

\subsection{Broader Impacts} % \Tao{repeats with limitation?}
Our work aims to develop a vision-based 3D object detection approach for roadside perception. The proposed method may produce inaccurate predictions, leading to incorrect decision-making for cooperative autonomous vehicles and potential traffic accidents. Furthermore, we propose a new perspective of leveraging height estimation to solve PV-BEV transformation, facilitating a high-performance and robust vision-centric BEV perception framework. Although considerable progress has been made with our proposed height net and height-based 2D-3D projection module, we believe it is worth further exploring how to combine height and depth estimations to extend to autonomous driving scenarios.

\subsection{Dynamic Discretization}
\begin{figure}[ht]
\centering	\includegraphics[width=8.5cm]{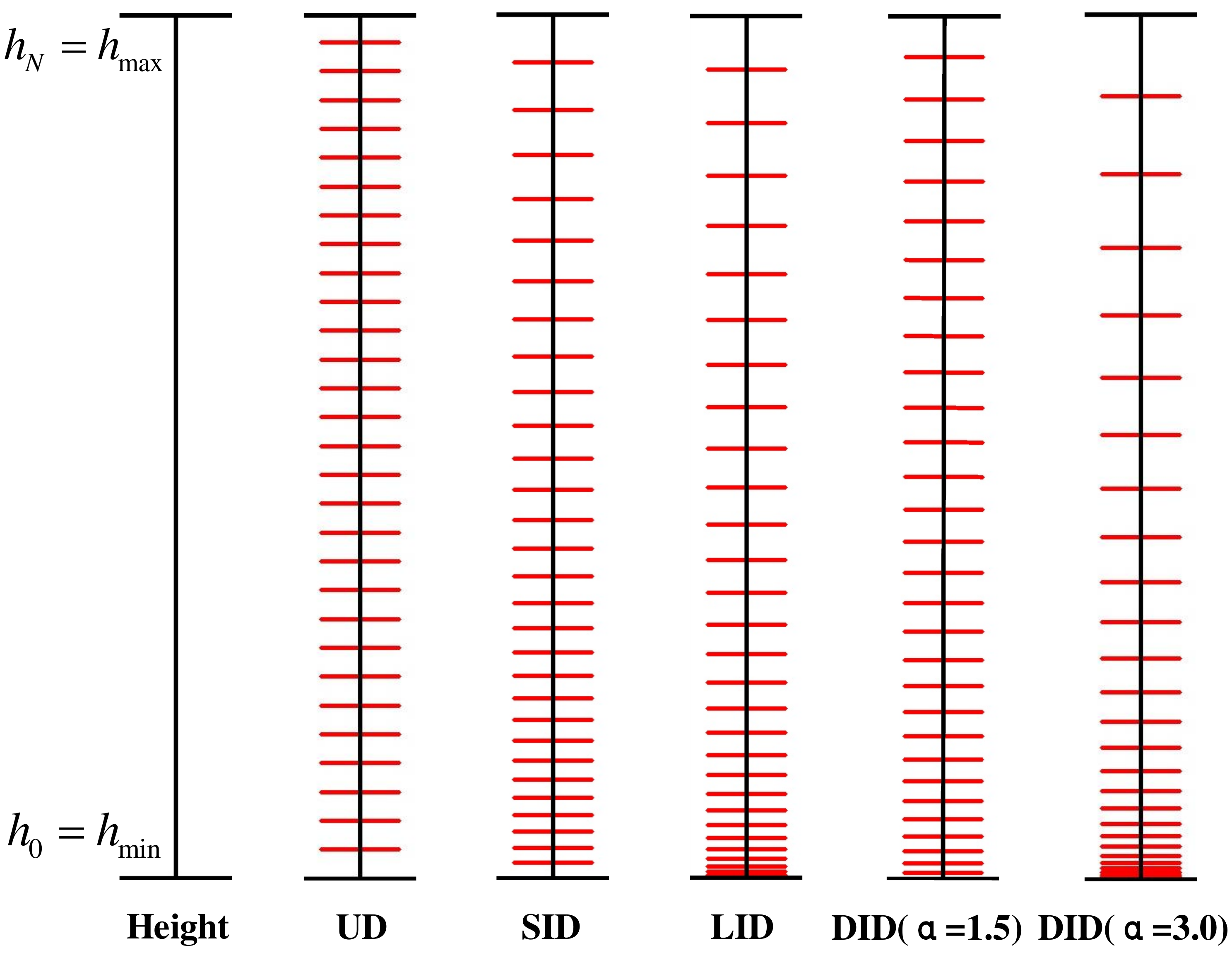}
	\caption{\textbf{Height Discretization Methods.} Height $h_i$ is discretized over a height range $[h_{min}, h_{max}]$ into $N$ discrete bins. From left to right, these are uniform discretization (UD), spacing-increasing discretization (SID), linear-increasing discretization (LID) and the adjustable dynamic-increasing discretization(DID). For the dynamic-increasing discretization (DID) strategy, height bins with large $\alpha$ are more densely distributed when approaching the $h_{min}$ than the small hyper-parameter $\alpha$ conditions.}
\label{fig:discretization}
\end{figure}

The height discretization can be performed with uniform discretization (UD) with a fixed bin size, spacing-increasing discretization (SID)~\cite{HuanFu2018DeepOR} with increasing bin sizes in logspace, linear-increasing discretization (LID)~\cite{YunleiTang2020Center3DCM}and our proposed dynamic-increasing discretization (DID) with adjustable bin sizes. The above four height discretization techniques are visualized in Fig. \ref{fig:discretization}. Following DID strategy, the distribution of height bins can be dynamically adjusted with different hyper-parameter $\alpha$.

%figure7

\subsection{Results on V2X-Sim Dataset}
To certify the effectiveness of our method in multi-view scenarios, we conduct experiments on V2X-Sim~\cite{li2022v2x} simulation dataset that contains four surround roadside cameras.
As shown in Tab.~\ref{v2x_sim_rebuttal}, our BEVHeight surpass the BEVDepth by more than 10.88\%, 21.15\% on vehicle and cyclist respectively, which verifies the effectiveness of our method.

\begin{table}[h!t]
 \scriptsize\centering\addtolength{\tabcolsep}{1.0pt}
\caption{\textbf{Comparison on the V2X-Sim Detection Benchmark.}}
 \begin{tabularx}{1.0\linewidth}{l|ccc|ccc}
 \toprule
 \multirow{3}{*}{Method} &
\multicolumn{3}{c|}{$\text{Vehicle}_{(IoU=0.5)}$} & \multicolumn{3}{c}{$\text{Cyclist}_{(IoU=0.25)}$} \\
 \cmidrule(r){2-7}
   & Easy & Mod. & Hard & Easy & Mod. & Hard  \\
 \midrule
 {BEVDepth~\cite{li2022bevdepth}}  &	81.99&	81.39&	81.31&	45.95&	45.93&	45.90\\
\rowcolor{cyan!30}{BEVHeight} &  92.80&	 92.27&	 92.15&	67.24& 67.08& 67.00\\
\bottomrule
\end{tabularx}
\label{v2x_sim_rebuttal}
\vspace{-0.3cm}
\end{table}

\subsection{Effectiveness on multi depth-based Detectors} We extend our modules on BEVDepth\cite{li2022bevdepth} and BEVDet~\cite{huang2021bevdet} on
 DAIR-V2X-I\cite{yu2022dair} and present the results here. Replacing the depth-based projection in BEVDepth\cite{li2022bevdepth}, our method achieves
a performance increase of 2.19\%, 5.87\%, 4.61\% on vehicle, pedestrian and cyclist. Similarly, our approach surpasses
the origin BEVDet by 8.56\%, 5.35\%, 8.60\% respectively.
% \begin{table*}[ht]
%  \centering\addtolength{\tabcolsep}{-0.6pt}
%  \resizebox{0.8\textwidth}{!}{
%  \begin{tabularx}{1.0\textwidth}{l|c|ccc|ccc|ccc}
%   \toprule
%  \multirow{3}{*}{Method} &  
%  \multirow{3}{*}{Modality}  
%  & \multicolumn{3}{c|}{$\text{Vehicle}_{(IoU=0.5)}$} & \multicolumn{3}{c|}{$\text{Pedestrian}_{(IoU=0.25)}$} & \multicolumn{3}{c}{$\text{Cyclist}_{(IoU=0.25)}$} \\
%     \cmidrule(r){3-11}
%      &  & Easy & Mid & Hard & Easy & Mid & Hard & Easy & Mid & Hard  \\
% \midrule

% PointPillars~\cite{lang2019pointpillars} & PointCloud &63.07 & 54.00 & 54.01 & 38.53 & 37.20 & 37.28 & 38.46 & 22.60 & 22.49 \\
% SECOND~\cite{yan2018second} & PointCloud &71.47 & 53.99 & 54.00 & 55.16 & 52.49 & 52.52 & 54.68 & 31.05 & 31.19 \\
% MVXNet~\cite{Sindagi2019MVX} & Image+PointCloud &71.04 & 53.71 & 53.76 & 55.83 & 54.45 & 54.40 & 54.05 & 30.79 & 31.06 \\
% \midrule
% ImvoxelNet~\cite{rukhovich2022imvoxelnet} &Image & 44.78 & 37.58 & 37.55 & 6.81 & 6.746 & 6.73 & 21.06 & 13.57 & 13.17 \\
% BEVFormer-R101$\ast$~\cite{li2022bevformer} & Image 	&	61.37&	50.73&	50.73&	16.89&	15.82&	15.95	&22.16&	22.13&	22.06\\
% BEVDepth-R101$\ast$~\cite{li2022bevdepth}&	Image 	&	76.01&	64.11&	64.18&	24.32&	24.96&	24.84	&46.45&	45.56&	45.69	\\

% \midrule
% BEVHeight-R101(Ours) & Image &	79.12&	67.95&	67.04&	29.85&	29.31&	29.07	&51.55&	51.39&	50.91\\
%     \bottomrule
%   \end{tabularx}
%   }
%   \caption{\textbf{Comparison on the DAIR-V2X-I val set.}}
%   \label{dair_sota}
% \end{table*}

\begin{table}[h!t]
 \scriptsize\centering\addtolength{\tabcolsep}{-4.9pt}
\caption{\textbf{Ablation studies on different depth-based methods.} Here, we conduct the evaluation on DAIR-V2X-I val set, and report the results of three types of objects, vehicle~(veh.), pedestrian~(ped.) and cyclist~(cyc.).}
 \begin{tabularx}{1.0\linewidth}{l|c|ccc|ccc|ccc}
 \toprule
 \multirow{3}{*}{Method} &
 \multirow{3}{*}{VT} &
\multicolumn{3}{c|}{$\text{Veh.}_{(IoU=0.5)}$} & \multicolumn{3}{c|}{$\text{Ped.}_{(IoU=0.25)}$} & \multicolumn{3}{c}{$\text{Cyc.}_{(IoU=0.25)}$} \\
 \cmidrule(r){3-11}
   &  & Easy & Mod. & Hard & Easy & Mod. & Hard & Easy & Mod. & Hard  \\
 \midrule
 
 \multirow{2}{*}{BEVDepth\cite{li2022bevdepth}} & D &	75.50&	63.58&	63.67&	34.95&	33.42&	33.27& 55.67&	55.47&	55.34 \\
  & {\cellcolor{cyan!30} H}&	{\cellcolor{cyan!30} 77.78}&	{\cellcolor{cyan!30} 65.77}&	{\cellcolor{cyan!30} 65.85}& {\cellcolor{cyan!30} 41.22}&	{\cellcolor{cyan!30} 39.29}&	{\cellcolor{cyan!30} 39.46}&  {\cellcolor{cyan!30} 60.23}&  {\cellcolor{cyan!30} 60.08}& {\cellcolor{cyan!30} 60.54} \\
\midrule
 \multirow{2}{*}{BEVDet~\cite{huang2021bevdet}} & D &  59.59& 	51.92&	51.81&  12.61& 12.43& 12.37& 34.91& 34.32& 34.21 	\\
& {\cellcolor{cyan!30} H}&	{\cellcolor{cyan!30} 69.42}&	{\cellcolor{cyan!30} 60.48}&	{\cellcolor{cyan!30} 59.68}& {\cellcolor{cyan!30} 18.11}&	{\cellcolor{cyan!30} 17.81}&	{\cellcolor{cyan!30} 17.74}&  {\cellcolor{cyan!30} 44.69}&  {\cellcolor{cyan!30} 42.92}& {\cellcolor{cyan!30} 42.34} \\
\bottomrule
\multicolumn{11}{l}{\scriptsize{VT denotes view transformation, D,H represents depth-based and height-based ones.}}
\end{tabularx}
\label{dair_rebuttal}
\vspace{-0.38cm}
\end{table}

% \subsection{More Results}
% \subsubsection{Results on Rope3D Dataset}
\subsection{More Results on DAIR-V2X-I Dataset}
Tab.~\ref{dair} shows the experimental results of deploying our proposed approach on the DAIR-V2X-I\cite{yu2022dair} val set. Under the same configurations (e.g., backbone and BEV resolution), our model outperforms the BEVDepth\cite{li2022bevdepth} baselines by a large marge, which demonstrates the admirable performance of our approach.

% \input{BEV-Height/latex/table/rope3d_het}

% 不同分辨率的泛化性
% 非重要内容，可放在补充材料中(分辨率非主要创新)
\begin{table*}[h!t]
 \small 
 \centering
 \addtolength{\tabcolsep}{1.4pt}
 \caption{\textbf{Comparison on the DAIR-V2X-I Detection Benchmark.} Here, we report the results of three types of objects: Vehicle, Pedestrian and Cyclist. Each object is categorized into three settings according to the difficulty defined in ~\cite{yu2022dair}. Our BEVHeight manages to surpass the BEVDepth baseeline over a margin of 2\% - 6\% under the same configurations.}
 \begin{tabularx}{1.0\textwidth}{l|cc|ccc|ccc|ccc}
  \toprule
 \multirow{4}{*}{Method} & \multicolumn{2}{c|}{\multirow{2.8}{*}{Scale of Detector}} & \multicolumn{9}{c}{AP3D}\\
    \cmidrule(r){4-12}
    & &  & \multicolumn{3}{c|}{$\text{Vehicle}_{(IoU=0.5)}$} & \multicolumn{3}{c|}{$\text{Pedestrian}_{(IoU=0.25)}$} & \multicolumn{3}{c}{$\text{Cyclist}_{(IoU=0.25)}$} \\
   \cmidrule(r){2-12}
   & Backbone & BEV & Easy & Middle & Hard & Easy & Middle & Hard & Easy & Middle & Hard\\  
    \midrule
    %  \midrule
     BEVDepth\cite{li2022bevdepth} &	R50&	128x128	&	73.05&	61.32&	61.19&	22.10	&21.57	&21.11	&42.85&	42.26&	42.09\\
    BEVDepth\cite{li2022bevdepth} &	R101&	128x128	&	74.81&	62.44&	62.31&	24.49	&23.33&	23.17&	44.93&	44.02&	43.84\\
    BEVDepth\cite{li2022bevdepth} &	R101&	256x256	&	75.50&	63.58&	63.67&	34.95&	33.42&	33.27& 55.67&	55.47&	55.34\\
    \midrule
    BEVHeight&	R50&	128x128	&	76.61	&64.71	&64.76&	27.34&	26.09&	26.33	&49.68&	48.84&	48.58\\
    BEVHeight&	R101&	128x128	&	76.93&	64.97&	65.03&	28.53&	27.15&	27.48& 51.39&	50.83	&50.44\\
    BEVHeight& R101&	256x256	&	\textbf{77.78}&	\textbf{65.77}&	\textbf{65.85}&	\textbf{41.22}&	\textbf{39.29}&	\textbf{39.46}	&\textbf{60.23}&	\textbf{60.08}&	\textbf{60.54}\\
  \bottomrule 
 % \multicolumn{21}{l}{Mid: Middle, Veh.: Vehicle, Ped.: Pedestrian, Cyc.: Cyclist.}
  \end{tabularx}
  \label{dair}
\end{table*}

\begin{figure*}[t!]
	\centering
	\includegraphics[width=\textwidth]{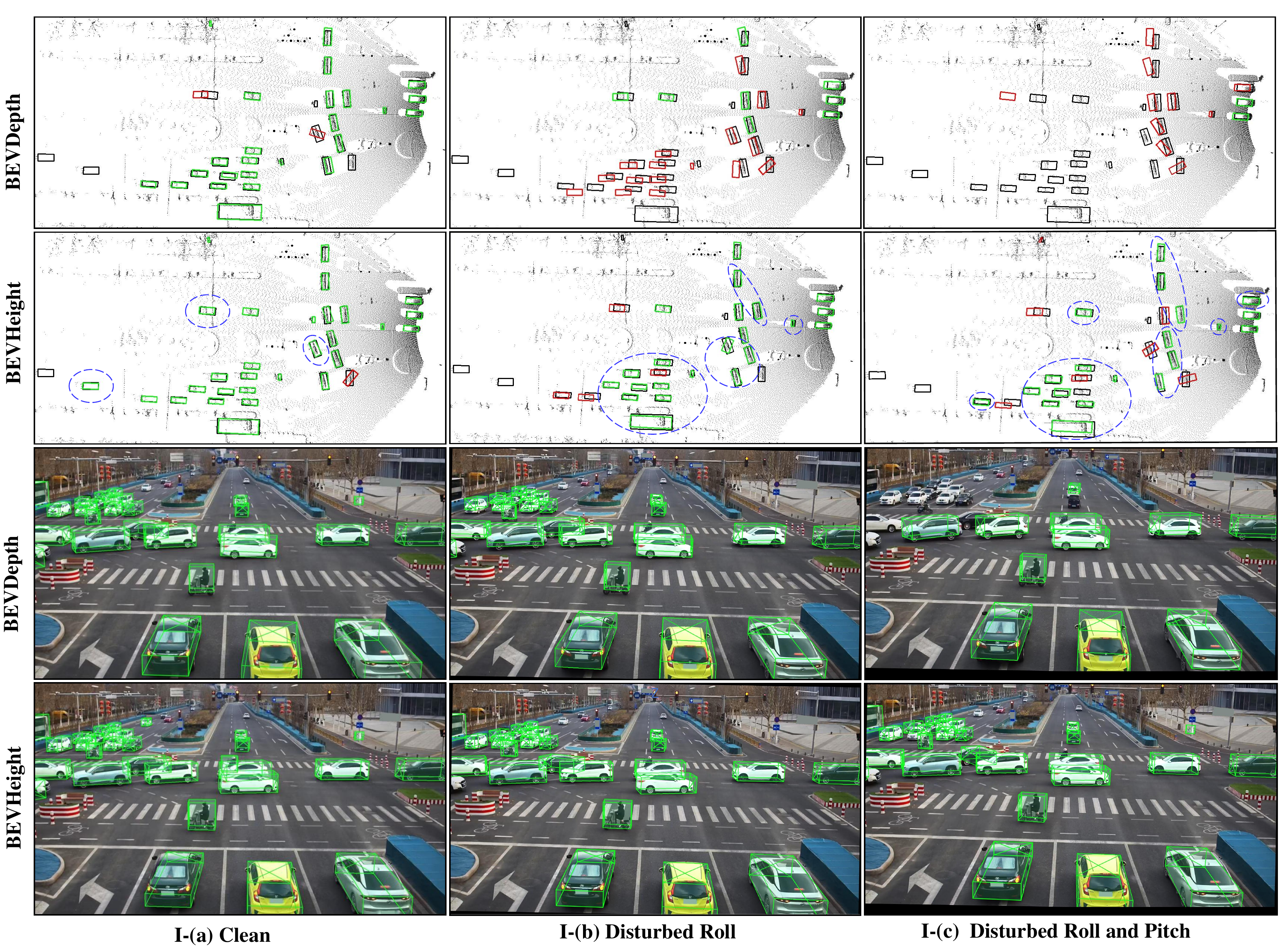}
	\caption{\textbf{Visualization Results of BEVDepth and our proposed BEVHeight under the extrinsic disturbance.} We use boxes in \textbf{{\color{red}red}} to represent false positives,  \textbf{{\color{green}green}} boxes for truth positives, and \textbf{{\color{black}black}} for the ground truth. The truth positives are defined as the predictions with IoU\textgreater 0.5 for vehicle and IoU\textgreater 0.25 for pedestrian and cyclist. I/II-(a) Clean means the original image without any processing; I/II-(b) Disturbed Roll denotes camera rotate 1 degree along roll direction; I/II-(c) Disturbed Roll and Pitch represents camera rotate 1 degree along roll and pitch directions simultaneously. We use \textbf{{\color{blue}blue}} dashed ovals to highlight the pronounced improvements in predictions.}
\label{fig:visualization_supp_1}
\vspace{-0.1cm}
\end{figure*}

\begin{figure*}[t]
	\centering
	\includegraphics[width=\textwidth]{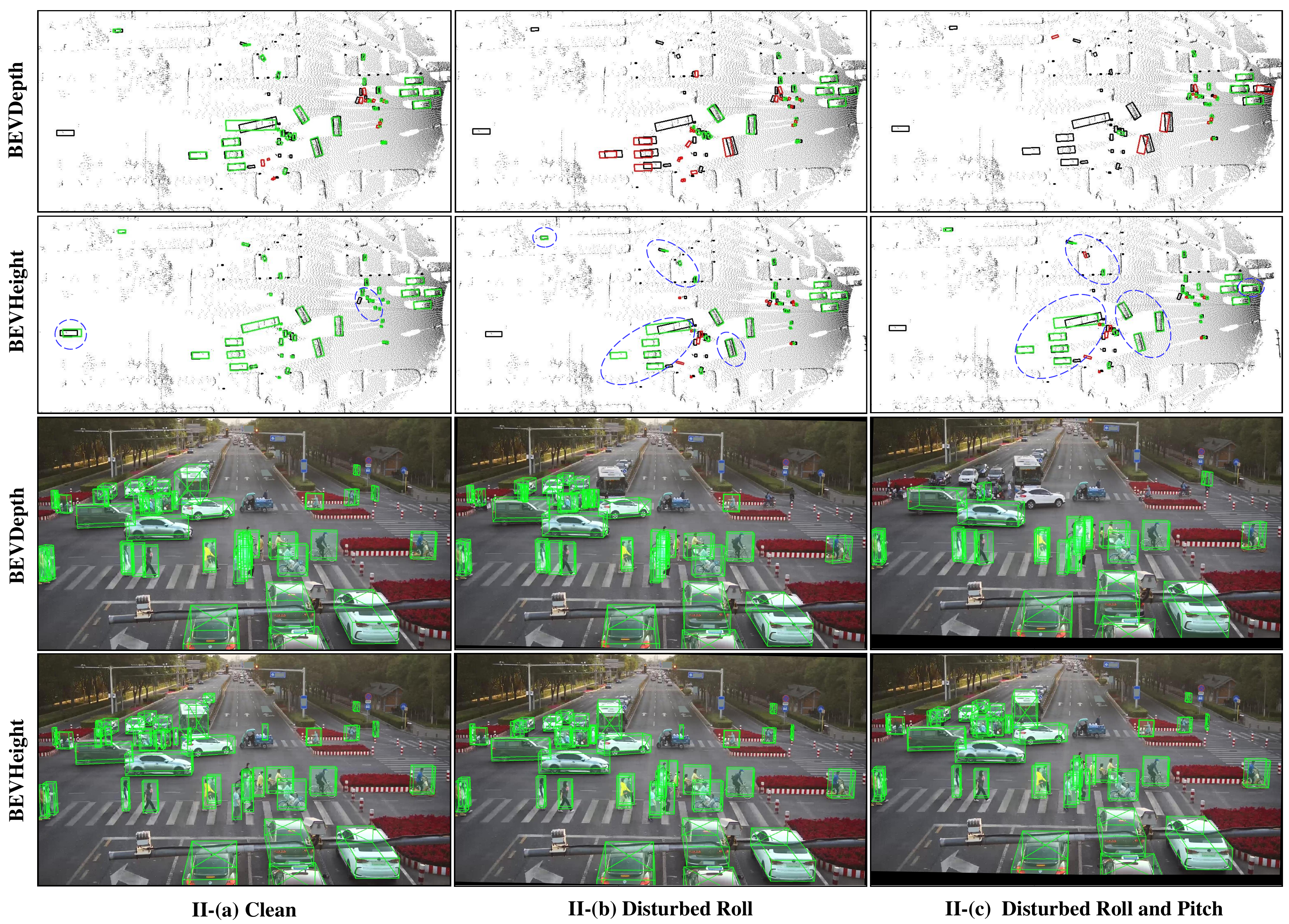}
	\caption{\textbf{Visualization Results of BEVDepth and our proposed BEVHeight under the extrinsic disturbance in another scene.}}
\label{fig:visualization_supp_2}
\vspace{-0.1cm}
\end{figure*}

\subsection{More Visualizations}
In Fig.~\ref{fig:visualization_supp_1} and Fig.~\ref{fig:visualization_supp_2}, we show more visualization results on the DAIR-V2X-I \cite{yu2022dair} dataset. As can be seen from the samples in I/II-(a) clean, our BEVHeight manage to detect objects in middle and long-distances. As for the  extrinsic disturbance cases in  I/II-(b) and I/II-(c),  our method can still guarantee the detection accuracy in terms of cars, pedestrian and cyclist. It can be concluded that our method can significantly improve the accuracy in middle and long-distances and the robustness to extrinsic disturbance.

% \Tao{Can we have some prediction results of height / depth?}
% \Lei{See Sec. ~\ref{sec:distance_error_analysis}, in process}

\end{appendices}

% \clearpage
% \input{BEV-Height/latex/5-appendix}

\end{document}